\documentclass[a4paper,fleqn,colorlinks,hidelinks]{cas-dc}

\usepackage[numbers]{natbib}
\usepackage{amsmath,amsfonts}
\usepackage{amssymb,amsthm}
\usepackage{mathtools}
\usepackage{bm}
\usepackage{booktabs}
\usepackage{multirow}
\usepackage{array}
\usepackage{enumitem}
\usepackage{tabularx}
\def\tsc#1{\csdef{#1}{\textsc{\lowercase{#1}}\xspace}}
\tsc{WGM}
\tsc{QE}
\tsc{EP}
\tsc{PMS}
\tsc{BEC}
\tsc{DE}

\begin{document}
\let\WriteBookmarks\relax
\def\floatpagepagefraction{1}
\def\textpagefraction{.001}
\shorttitle{Spatio-temporal Evolving Structural Representation of Action Units for Micro-expression Recognition}
\shortauthors{N. Sharma et~al.}

\title [mode = title]{STAG: Spatio-temporal Evolving Structural Representation of Action Units for Micro-expression Recognition}                      

\author[1]{Nandani Sharma}[type=editor,
                       auid=000,bioid=1,
                       orcid=0000-0003-0096-9754]
\cormark[1]
\ead{d22180@students.iitmandi.ac.in}
\ead[url]{https://sites.google.com/view/nandanisharma/home/}

\author[1,2]{Varun Sharma}
\cormark[1]
\ead{varun.240102085@iiitbh.ac.in}

\author[1]{Dinesh Singh}[type=editor,
                        orcid=0000-0001-8889-9847]

\affiliation[1]{organization={Vision Intelligence and Machine Learning (VIML) Group, School of Computing and Electrical Engineering, Indian Institute of Technology Mandi},
                city={Mandi},
                postcode={175005},
                state={Himachal Pradesh},
                country={India}}

\affiliation[2]{organization={Indian Institute of Information Technology Bhagalpur},
                addressline={Sabour Rd},
                city={Bhagalpur},
                postcode={813210},
                state={Bihar},
                country={India}}
\cortext[cor1]{Corresponding author}

\begin{abstract}
Micro-expression recognition is challenging due to extremely subtle and short-lived facial muscle
movements. Existing approaches depend heavily on apex–onset frames and often ignore fine-grained
inter-frame dynamics. They also treat spatial and temporal information separately, failing to jointly
model “where” facial actions occur and “when” they evolve. Fixed ROI connectivity and loosely
fused action unit (AU) cues further limit structured temporal alignment and reduce cross-dataset
generalization. As a result, the learned representations become highly specialized to the training
dataset and fail to generalize when tested on datasets containing different recording conditions,
ethnicities, camera setups, illumination variations, frame rates, and expression distributions. To
address these limitations, we propose STAG, a dynamic ROI–AU-coupled spatial–temporal network
that jointly models motion flow and adaptive facial connectivity. The framework first extracts optical
flow from discriminative frames using magnitude-based selection and temporal attention to capture
the most informative motion patterns. A dual-branch architecture is then employed, consisting of an
enhanced graph attention network for structured spatial reasoning over facial regions and a transformer
encoder for full-sequence temporal modeling. A bidirectional cross-attention module enables mutual
refinement between spatial and temporal features, tightly coupling motion flow with evolving ROI
relationships. Additionally, AU-guided dynamic connectivity allows facial region interactions to adapt
to muscle activation patterns. The transformer preserves subtle inter-frame dynamics missed by
apex-based approaches, improving semantic consistency and interpretability for XAI-based micro-
expression recognition. The fused representation is optimized using focal loss and evaluated on
benchmark datasets: CASME II, 4DME, DFME, NaME, SAMM, and SMIC-HS. Extensive qualitative
and quantitative experiments demonstrate the robustness, generalization capability, interpretability,
and computational efficiency of the proposed framework. The results confirm the effectiveness of
adaptive relational reasoning, AU-guided dynamic connectivity, and deep spatial–temporal feature
fusion for accurate and explainable micro-expression recognition across diverse datasets.
\end{abstract}




\maketitle

\section{Introduction}
Micro-expressions are involuntary, extremely brief fa-
cial muscle movements that reveal hidden emotions~\cite{zhang2025hfa}.
Despite their importance in deception detection, security,
healthcare, and human–computer interaction, 
micro-expression
recognition (MER) remains highly challenging due to their
low intensity, short duration, localized facial activations, and
limited annotated data~\cite{malik2026comprehensive,shangguan2025facial}. These characteristics make
robust and generalized MER difficult, requiring models that
can effectively capture subtle spatial-temporal dynamics.

With deep learning, CNN, RNN, and 3D CNN-based
methods~\cite{shukla2022micro,wang2021micro,irawan2023spontaneous} and later hybrid and transformer-based
models~\cite{wang2024two,ma2025multi,zhang2025hfa} significantly improved performance.
However, CNN-based approaches have limited temporal re-
ceptive fields and weak interpretability, while transformer-
based methods~\cite{vaswani2017attention,wang2025multi,liu2025mer,sharma2026improved} better capture long-range dependencies but often decouple spatial and temporal modeling.

Graph-based MER methods~\cite{lei2021micro,zhao2021sta,zhang2023adaptive,fang2025graph,sharma2025x, sharma2025exp} further model
facial regions as structured ROIs using GNNs, improving
relational reasoning and interpretability. However, most approaches rely on static adjacency matrices, loosely integrate
AU cues~\cite{xie2020assisted,lei2021micro,WeiPLLYZ24}, and independently optimize spatial and
temporal learning, limiting their ability to model dynamic
facial interactions and generalize across datasets.

To overcome the limitations of local convolutional mod-
eling, transformer-based MER frameworks have recently
attracted increasing attention due to their ability to capture
long-range temporal dependencies through self-attention
mechanismss~\cite{vaswani2017attention}. MMTNet~\cite{wang2025multi} introduced multi-modal
multi-scale transformer learning for joint motion and appear-
ance modeling, while HLoRA-MER~\cite{shao2025high} utilized parameter-
efficient LoRA fine-tuning over DinoV2 features to improve
subtle facial representation learning with reduced compu-
tational cost. MER-CLIP~\cite{liu2025mer} further integrated visual-
language alignment by associating action unit (AU) descriptions with visual facial dynamics. In addition, MOL~\cite{shao2025mol}
jointly optimized MER, optical flow estimation, and land-
mark detection within a unified transformer-graph-style
convolutional framework. More recently, SODA4MER~\cite{zhang2025dynamic}
proposed an apex-free self-supervised framework based
on oriented deformation estimation, while CausalNet~\cite{zhang2025rethinking}
improved robustness against key-frame annotation errors
using causal motion reasoning. ME-NAS further explored
neural architecture search for automated MER network
optimization. Despite these advances, many transformer-
based methods still treat spatial structure modeling and
temporal sequence learning independently without explicitly
coupling facial region interactions with temporal evolution.

To explicitly model relationships among facial regions,
graph neural networks (GNNs) have emerged as an effec-
tive paradigm for MER. By representing facial landmarks
or regions of interest (ROIs) as graph nodes and their
anatomical relationships as edges, graph-based methods
can perform structured relational reasoning among facial
muscle movements~\cite{kipf2017gcn}. Early graph-based MER methods
such as AU-GCN~\cite{lei2021micro} and STA-GCN~\cite{zhao2021sta} incorporated
facial action units into graph learning to capture local
muscle interactions. Subsequently, GCL~\cite{lao2022temporal} introduced
graph contrastive learning to model temporal variation in
ME sequences, while AGT~\cite{zhang2023adaptive} dynamically constructed
adaptive ROI graphs through graph attention mechanisms.
ATM-GCN~\cite{zhang2024adaptive} further improved clip-level temporal motion aggregation using adaptive graph convolution. More
recent graph-based approaches such as SGCN~\cite{TangC24}, GTS-GN+AU~\cite{WeiPLLYZ24}, OFVIG-Net~\cite{DBLP:journals/ijon/ZhangZSTWL25}, GTA~\cite{wang2025gta}, ADSS~\cite{DBLP:journals/tbbis/0002BW0L025}, DGAT~\cite{wang2026micro}, and AU-GCN-CUR~\cite{fang2025graph} enhanced graph
learning through stochastic graph structures, geometric graph
reasoning, transformer integration, and adaptive discriminative region selection. These methods demonstrated that
graph reasoning can effectively capture structured spatial dependencies among facial regions and improve interpretability.

Nevertheless, existing graph-based MER methods still
face several fundamental challenges. First, most methods
rely on predefined or static adjacency matrices derived from
facial geometry priors, which fail to capture dynamically
evolving muscle interactions during micro-expression pro-
gression. Second, AU information is often integrated in a
loosely coupled manner without temporally aligned rela-
tional reasoning. Third, spatial graph learning and temporal
sequence modeling are commonly optimized independently,
preventing mutual refinement between \emph{where} facial motions
occur and \emph{when} they evolve over time. Consequently, ex-
isting methods struggle to jointly model adaptive spatial
connectivity and fine-grained temporal evolution in a unified
framework.

The facial action coding system (FACS)~\cite{ekman1978facs} provides
valuable physiological priors for understanding facial ex-
pressions through AUs. Although several
MER methods incorporate AU cues~\cite{xie2020assisted,lei2021micro,WeiPLLYZ24}, they gen-
erally fail to exploit dynamic AU-driven connectivity evolu-
tion across temporal sequences. Furthermore, most existing
approaches exhibit limited cross-dataset generalization due
to overfitting on small-scale MER datasets and insufficient
modeling of adaptive relational dependencies.

To address these limitations, we propose STAG (\textbf{S}patio-
\textbf{T}emporal Evolving \textbf{S}tructural Representation of \textbf{A}ction Units for
Micro-expression Reco\textbf{G}nition), a unified spatial–temporal
framework that jointly models motion flow and adaptive
ROI connectivity. It extracts discriminative optical flow
using magnitude-based selection and temporal attention,
followed by dual-branch graph attention and transformer-based modeling. A bidirectional cross-attention mechanism
tightly couples spatial and temporal features, while AU-guided dynamic graphs enable evolving facial connectivity.
This unified design improves interpretability, captures fine-grained motion evolution, and enhances cross-dataset generalization for robust MER. 

To tightly couple spatial and temporal representations,
we introduce a bidirectional cross-attention module that en-
ables mutual refinement between graph-based ROI features
and temporal motion representations. This design allows
the framework to simultaneously learn \emph{where} subtle facial
motions occur and \emph{when} they evolve across the full sequence.
Furthermore, dynamic ROI connectivity is adaptively modu-
lated through AU-guided attention, allowing the graph struc-
ture to evolve according to facial muscle activation patterns
instead of relying on fixed adjacency matrices. Compared
with existing graph-based MER methods, the proposed dy-
namic AU-guided connectivity provides stronger semantic
consistency, improved interpretability, and enhanced ex-
plainability for explainable MER (XAI-MER). Extensive
experiments on six benchmark datasets (CASME II~\cite{yan2014casme2}, 4DME~\cite{li20224dme}, DFME~\cite{zhao2023dfme}, NaME~\cite{liu2025name}, SAMM~\cite{davison2018samm}, SMIC-HS~\cite{li2013smic}) under LOSO, Group K-Fold, and Stratified K-
Fold protocols demonstrate that STAG achieves state-of-
the-art (SOTA) performance, robustness, and strong cross-dataset
generalization. The main contributions of this
work are summarized as follows:
\begin{itemize}
\item  We propose STAG, a unified dynamic ROI–AU-
coupled spatial-temporal framework for MER that
jointly models adaptive facial region connectivity
and full-sequence motion evolution within a single
learning architecture.
\item We design an enhanced graph attention network (E-
GAT) as the core spatial reasoning module, which
dynamically captures AU-guided facial region de-
pendencies and enables adaptive graph learning that
evolves with muscle activation patterns.
\item STAG uses a bidirectional cross-attention mechanism
between the enhanced GAT and a transformer-based
temporal encoder, enabling fine-grained mutual re-
finement of spatial and temporal representations while
preserving subtle inter-frame motion dynamics.
\end{itemize}

The remainder of this paper is organized as follows.
Section~\ref{sec:RW} reviews related work and discusses SOTA 
methods in micro-expression recognition. Section~\ref{sec:Method} presents
the proposed STAG framework and its methodology. Section~\ref{sec:ES} describes the experimental setup and reports the results along with comprehensive analyses. Finally, section~\ref{sec:Con} concludes the paper and
outlines potential directions for future work.
\section{Related Work}\label{sec:RW}
Early MER methods primarily relied on optical-flow-
guided CNN architectures to capture subtle facial motion
patterns. OFF-ApexNet employed handcrafted optical-flow
features and GAN-based augmentation to improve motion
representation learning~\cite{liong2020evaluation}, while Graph-TCN incorporated
motion magnification and temporal convolution to enhance
local motion perception~\cite{lei2020novel}. DSTAN and ConvLSTM fur-
ther explored dual-stream attention mechanisms and recur-
rent temporal modeling for spatio-temporal feature extrac-
tion~\cite{wang2021micro,shukla2022micro}. Although these approaches achieved promising
performance, they primarily depend on onset-apex repre-
sentations and local receptive fields, limiting their ability
to capture complete temporal evolution and long-range de-
pendencies throughout the micro-expression sequence. Fur-
thermore, the learned CNN features often lack physiological
interpretability because facial muscle interactions are not
explicitly modeled. 

To address these limitations, the proposed STAG frame-
work leverages a transformer encoder to model the entire
temporal sequence rather than relying solely on apex-based
information. By capturing long-range temporal dependen-
cies and integrating an E-GAT for ROI-level relational rea-
soning, STAG learns more discriminative and interpretable
spatio-temporal representations.

Subsequent research focused on multi-scale feature ex-
traction and dual-branch fusion strategies. LRCN-STAN
integrated CBAM and LSTM modules for attentive spatio-
temporal learning, while TFT combined CNN and Swin
Transformer features to capture hierarchical facial repre-
sentations~\cite{zhang2026lrcn,wang2024two}. DSBICNet, DBDE-Net, and SFML-Net
further enhanced motion-sensitive feature learning through
dual-stream architectures and local-global feature refine-
ment~\cite{shou2026dsbicnet,wang2026dbde,zhang2026sfml}. Similarly, MARNet, FDP, MADV-Net,
and MDMO improved MER performance through motion
enhancement, dynamic ranking, self-supervised learning,
and efficient optical-flow estimation~\cite{xu2026motion,shao2025micro,kong20253d,zhang2025fast}. Addi-
tional methods such as SCFRAM, MPFNeT, DLRRF-MER,
3DCNN, Concat-CNN, and LSSNet investigated supervised
contrastive learning, multi-prior feature fusion, long-video
spotting, and lightweight feature extraction strategies~\cite{li2026micro,ma2025multi,islam2023highly,irawan2023spontaneous,yang2023deep,yu2021lssnet}. Despite their effectiveness, most of these
methods learn spatial and temporal information indepen-
dently and perform feature fusion only at later stages, which
restricts effective interaction between facial spatial struc-
tures and temporal dynamics.

In contrast, STAG introduces a bidirectional graph-
transformer cross-attention mechanism that enables con-
tinuous information exchange between graph-based spatial
reasoning and transformer-based temporal modeling. Com-
bined with dynamic ROI-AU graph construction, tempo-
rally smoothed adjacency updates, and AU-guided dynamic
connectivity, the proposed framework learns adaptive facial
relationships and unified spatio-temporal representations,
effectively overcoming the limitations of existing CNN-
based and dual-branch MER approaches.
\subsection{Graph-based MER}
Graph-based MER methods model facial muscle de-
pendencies and AU relationships using graph convolution
and relational learning. AU-GCN pioneered facial graph
representation learning by integrating landmark-aware node
learning and AU fusion~\cite{lei2021micro}. However, graph connectivity
remains largely predefined and static, limiting adaptation
to changing facial muscle interactions, STA-GCN models
spatio-temporal AU dependencies but still relies on pre-
defined graph structures and cannot dynamically update
facial relationships during expression evolution~\cite{zhao2021sta}. GCL
introduced graph contrastive learning for temporal variation
modeling~\cite{lao2022temporal}. However, graph connectivity remains weakly
adaptive and does not explicitly incorporate evolving AU
relationships. AGT introduces adaptive graph attention but
still lacks tight temporal coupling and AU-guided dynamic
evolution~\cite{zhang2023adaptive}. Although these methods improved structured
facial representation learning, most rely on pre-defined or
weakly adaptive graph connectivity.

Recent graph-based frameworks focused on dynamic
graph construction and temporal dependency modeling.
ATM-GCN improves temporal motion aggregation but spa-
tial and temporal learning remain relatively independent~\cite{zhang2024adaptive},
while SGCN improves generalization through stochastic
graph learning but lacks physiologically meaningful AU-
driven connectivity evolution~\cite{TangC24}. Although AU informa-
tion is included, the AU integration remains loosely cou-
pled and not temporally aligned~\cite{WeiPLLYZ24}. OFVIG-Net directly
learned graph representations from optical-flow patches
using Vision-GNN~\cite{DBLP:journals/ijon/ZhangZSTWL25}, whereas GTA combined graph convolution and transformers for AU-aware relational modeling~\cite{wang2025gta}. ADSS selected discriminative graph regions adaptively
for subtle motion representation~\cite{DBLP:journals/tbbis/0002BW0L025}. More recently, SOFP
exploited bilateral facial symmetry and region-aware attention fusion for robust local-global motion learning~\cite{yu2026sofp},
while AU-GCN-CUR extended graph reasoning for digital-
human micro-expression rendering~\cite{fang2025graph}.

These methods improve graph learning using Vision GNNs,
transformers, or discriminative region selection. However,
most still depend on static or weakly adaptive graph structures and fail to jointly optimize spatial-temporal interaction.
To overcome the limitations of existing graph-based MER
methods, STAG introduces a dynamic ROI-AU graph construction mechanism that adaptively models facial muscle
interactions. The framework further employs temporally
smoothed adjacency updates and an E-GAT to capture
evolving spatial relationships. In addition, a bidirectional
graph-transformer cross-attention module enables effec-
tive integration of spatial and temporal information, while
AU-guided dynamic connectivity ensures physiologically
meaningful feature learning. Together, these components
provide adaptive relational reasoning and robust spatio-
temporal representation learning, effectively addressing the
shortcomings of prior graph-based approaches.






\begin{table*}[t]
\centering
\caption{Comparison of representative graph-based, transformer-based, and AU-guided micro-expression recognition methods with the proposed STAG framework. ``Partial'' indicates that the capability is only indirectly or partially incorporated.}
\label{tab:method_comparison}

\resizebox{\textwidth}{!}{
\begin{tabular}{lcccccccccccc}
\hline
\textbf{Component}
& \textbf{AU-GCN}
& \textbf{STA-GCN}
& \textbf{GCL}
& \textbf{AGT}
& \textbf{ATM-GCN}
& \textbf{GTS+GN+AU}
& \textbf{OFVIG-Net}
& \textbf{GTA}
& \textbf{DGAT}
& \textbf{MER-CLIP}
& \textbf{MOL}
& \textbf{STAG}
\\
\hline

Graph-based Spatial Modeling
& \checkmark
& \checkmark
& \checkmark
& \checkmark
& \checkmark
& \checkmark
& \checkmark
& \checkmark
& \checkmark
& $\times$
& \checkmark
& \checkmark
\\

Dynamic Graph Construction
& $\times$
& $\times$
& Partial
& \checkmark
& Partial
& Partial
& Partial
& \checkmark
& \checkmark
& $\times$
& Partial
& \checkmark
\\

AU Guidance
& \checkmark
& \checkmark
& $\times$
& $\times$
& $\times$
& \checkmark
& $\times$
& \checkmark
& Partial
& \checkmark
& $\times$
& \checkmark
\\

Transformer Encoder
& $\times$
& $\times$
& $\times$
& \checkmark
& $\times$
& $\times$
& $\times$
& \checkmark
& \checkmark
& \checkmark
& \checkmark
& \checkmark
\\

Temporal Sequence Modeling
& Partial
& \checkmark
& \checkmark
& Partial
& \checkmark
& Partial
& Partial
& \checkmark
& \checkmark
& \checkmark
& \checkmark
& \checkmark
\\

Graph--Transformer Fusion
& $\times$
& $\times$
& $\times$
& Partial
& $\times$
& $\times$
& $\times$
& \checkmark
& \checkmark
& $\times$
& Partial
& \checkmark
\\

Bidirectional Cross-Attention
& $\times$
& $\times$
& $\times$
& $\times$
& $\times$
& $\times$
& $\times$
& $\times$
& $\times$
& $\times$
& $\times$
& \checkmark
\\

AU-driven Adaptive Connectivity
& Partial
& Partial
& $\times$
& $\times$
& $\times$
& Partial
& $\times$
& Partial
& Partial
& \checkmark
& $\times$
& \checkmark
\\

Temporal Graph Evolution
& $\times$
& Partial
& Partial
& $\times$
& Partial
& $\times$
& $\times$
& Partial
& $\times$
& $\times$
& $\times$
& \checkmark
\\

Full-Sequence Modeling (Beyond Apex)
& $\times$
& Partial
& Partial
& Partial
& Partial
& Partial
& Partial
& \checkmark
& \checkmark
& \checkmark
& \checkmark
& \checkmark
\\

Explainability-Oriented Design
& \checkmark
& \checkmark
& Partial
& \checkmark
& Partial
& \checkmark
& Partial
& Partial
& Partial
& \checkmark
& Partial
& \checkmark
\\

Cross-Dataset Generalization Focus
& $\times$
& $\times$
& $\times$
& $\times$
& $\times$
& $\times$
& $\times$
& $\times$
& Partial
& \checkmark
& Partial
& \checkmark
\\

\hline
\end{tabular}
}
\end{table*}
\subsection{Transformer and Foundation-Model-based MER}
Transformer-based MER frameworks have demonstrated
superior capability in modeling long-range spatio-temporal
dependencies. Captures hierarchical global dependencies
but does not explicitly model dynamic facial region inter-
actions~\cite{zhang2025hfa}, while MMTNet employed multi-modal multi-
scale transformers with cross-modal contrastive learning
but treats spatial structure and temporal evolution sepa-
rately~\cite{wang2025multi}. HLoRA-MER and DGAT leveraged DinoV2 and
parameter-efficient LoRA fine-tuning to improves efficiency
through LoRA fine-tuning but remains dependent on ap-
pearance features and apex-centered learning~\cite{shao2025high,wang2026micro}. Al-
though these methods achieve strong representation learning
performance, most still rely heavily on apex-based motion
extraction and weak graph-temporal interaction.

Several recent approaches integrated self-supervised
learning, causal reasoning, and multimodal learning. MER-
CLIP aligned AU-guided textual semantics with visual
motion representations through CLIP-based cross-modal
learning but does not explicitly learn evolving graph struc-
tures~\cite{liu2025mer}. MOL jointly optimized MER, optical flow es-
timation, and landmark detection using transformer-graph-
style convolution but lacks explicit dynamic graph-temporal
interaction~\cite{shao2025mol}. SODA4MER proposed an apex-free self-
supervised deformation learning framework but does not in-
corporate structured graph reasoning~\cite{zhang2025dynamic}, while CausalNet
improved robustness against key-frame annotation errors
through causal motion reasoning but lacks dynamic ROI
connectivity modeling~\cite{zhang2025rethinking}. ME-NAS introduced a neural
architecture search strategy specialized for MER~\cite{verma2026me}.

In summary, existing CNN, graph, and transformer-
based MER methods have achieved promising performance;
however, they often model spatial and temporal information
independently, rely on static ROI relationships, or employ
apex-centered representations that overlook subtle inter-
frame dynamics. Moreover, the interaction between facial
muscle activations, spatial dependencies, and temporal evo-
lution remains insufficiently explored.
To address these limitations, the proposed STAG framework introduces a unified and interpretable architecture that
jointly models dynamic ROI connectivity, AU-guided relational reasoning, and full-sequence temporal
evolution. By integrating graph-based spatial learning with
transformer-based temporal encoding through bidirectional
cross-attention, STAG enables mutual refinement between
‘where” and ‘when” micro-expressions occur. Furthermore,
AU-modulated graph attention dynamically adapts facial
relationships according to underlying muscle activations,
enhancing both representation learning and interpretability.
Consequently, STAG provides more discriminative spatial-temporal features, improved cross-domain generalization,
and stronger explainability for robust micro-expression recognition across diverse benchmark datasets.

Table \ref{tab:method_comparison} summarizes the methodological characteristics
of representative graph-based, transformer-based, and AU-
guided micro-expression recognition approaches. While ex-
isting methods incorporate subsets of these capabilities,
none simultaneously integrate dynamic graph construction,
temporal graph evolution, bidirectional cross-attention, AU-
driven adaptive connectivity, and unified ROI–AU coupled
learning. The proposed STAG framework is designed to
bridge these gaps by jointly modeling evolving spatial re-
lationships and temporal dynamics within a unified graph-
transformer architecture.

\begin{figure*}[!t]
    \centering
    \includegraphics[width=1.0\linewidth, keepaspectratio]{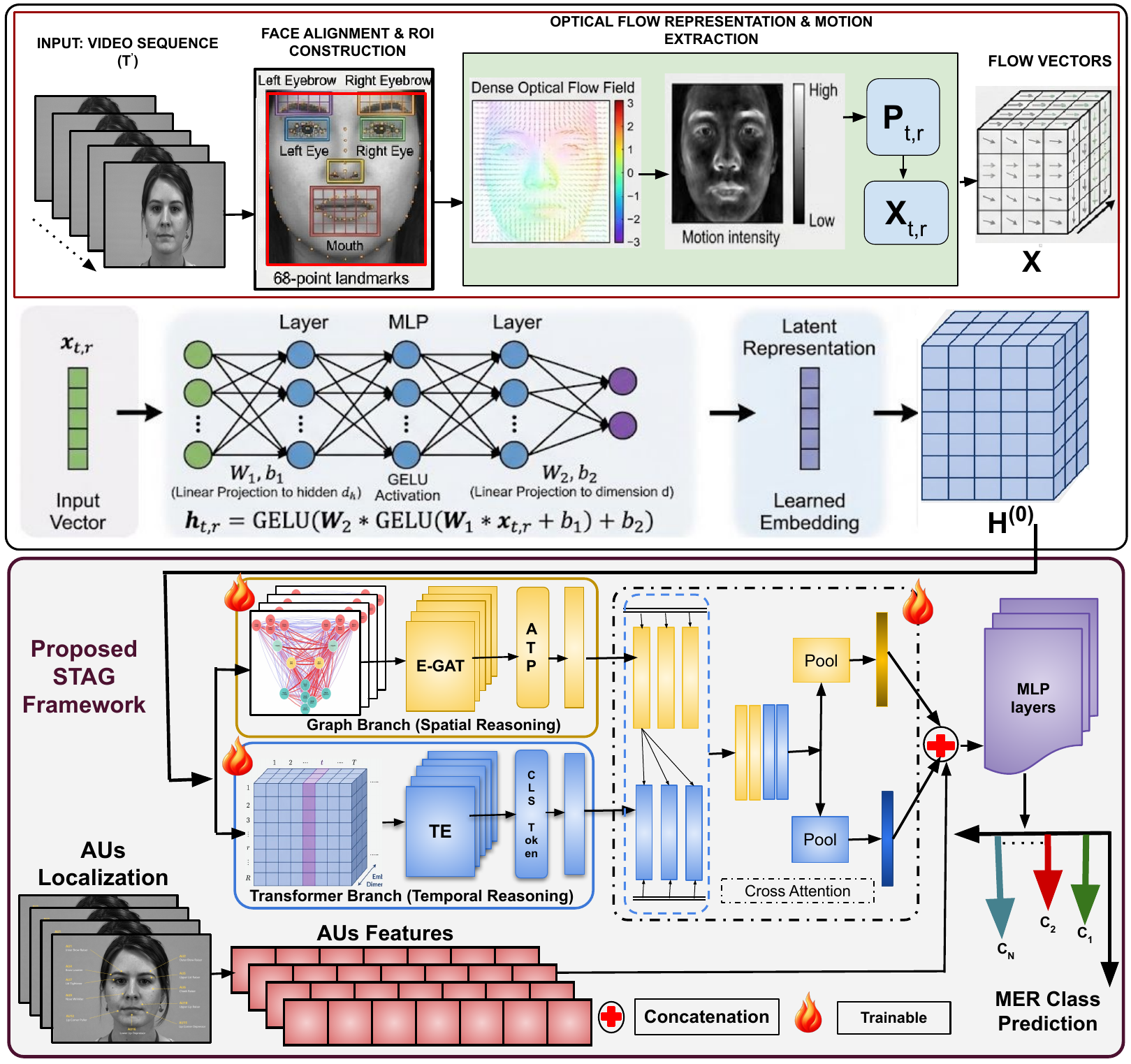}
    \caption{Overview of the proposed STAG framework for MER. Given an input facial video sequence, facial landmarks are detected and ROIs are constructed around key facial components, including the eyes, eyebrows, and mouth. Dense optical flow is then computed to capture subtle facial motions, producing motion intensity maps and flow vectors represented as feature tensor $\mathbf{X}$. These features are projected into a latent embedding space through a MLP, yielding the initial representation $\mathbf{H}^{(0)}$. Simultaneously, facial AUs are localized and corresponding AU features are extracted. The proposed STAG architecture consists of a graph branch for spatial reasoning using E-GAT and a transformer branch for temporal reasoning. The outputs of both branches are integrated through cross-attention and pooling operations to learn complementary spatial-temporal representations. The fused features are further concatenated with AU features and passed through trainable MLP layers to perform micro-expression classification, producing the final expression class predictions. Flame icons denote trainable modules within the framework.}
    \label{fig:architecture}
  \end{figure*}
\section{Methodology}\label{sec:Method}
The proposed framework processes a micro-expression
video through a unified pipeline consisting of geometric fea-
ture extraction, optical flow modeling, graph-based spatial
reasoning, temporal transformer modeling, and cross-modal
fusion. Given an input video, the model first converts raw
frames into structured motion representations, which are
then encoded using a joint Graph-Transformer architecture
with AUs conditioning for final classification (shown in the Figure~\ref{fig:architecture}). 
\subsection{Problem Formulation}
Let $\mathbf{X} \in \mathbb{R}^{T \times R \times 2}$ denote a temporally-standardized opti-
cal flow sequence, where $T$ is the fixed number of temporal
frames, $R$ is the number of facial ROIs, and the trailing
dimension of $2$ captures the horizontal and vertical flow
components $(u_{t,r}, v_{t,r})$ for frame $t$ and ROI $r$. The goal is to
predict the micro-expression class $\hat{y} \in \mathcal{Y}$ via:
\begin{equation}
\hat{y} = \arg\max_{c \in \mathcal{Y}}\; f_\theta\left(\mathbf{X},\,\mathbf{v}_\mathrm{au}\right),
\end{equation}
where $f_\theta(\cdot)$ is the proposed network parameterized by the set of all trainable parameters $\theta$, $c$ indexes over the set of micro-expression classes $\mathcal{Y}$, and $\mathbf{v}_\mathrm{au} \in \{0,1\}^{N_\mathrm{AU}}$ is the optional binary AU indicator vector of length $N_\mathrm{AU}$. 

\subsection{Face Alignment and ROI Construction}
A micro-expression video is represented as a sequence of frames $\mathcal{V} = \{ I_t\}_{t=1}^{T}$, where each frame $I_t \in \mathbb{R}^{H \times W \times 3}$ is an RGB image. 
A face-centred crop of size 256 x 256 is extracted using the eye-corner
landmarks (points 39 and 42) to define scale and center.
Six anatomical ROI bounding boxes are derived from landmark subsets:
left eyebrow (points 17--21), right eyebrow (22--26), left eye (36--41),
right eye (42--47), nose tip (30--35), and mouth (48--67).
Within each ROI a set of key landmark positions defines finer sub-windows,
giving $R\!=\!18$ motion channels in total
(4 left-eyebrow, 4 right-eyebrow, 6 mouth, 2 nose-tip, 1 left-eye, 1 right-eye)~\cite{DBLP:journals/taffco/WeiPLLYZ24, li2026micro}. For the initial frame, a frontal-face detector and a 68-point facial landmark predictor~\cite{dlib09} are applied to extract a set of 2D facial keypoints, and landmarks detected in the first frame are tracked across the sequence:
\begin{equation}
\mathcal{L}_t = \left\{ \mathbf{l}_t^{(i)} \right\}_{i=1}^{68}, \quad \mathbf{l}_t^{(i)} = \left(x_t^{(i)}, y_t^{(i)}\right) \in \mathbb{R}^{2}.
\end{equation}
To minimize inter-subject variation and ensure scale invari-
ance across subjects, we compute the inter-ocular distance
using predefined eye-corner landmarks indexed by $a$ and $b$:
\begin{equation}
d_e = \left\| \mathbf{l}_t^{(a)} - \mathbf{l}_t^{(b)} \right\|_2.
\end{equation}
The normalization scaling coefficient is defined as $g = \frac{d_e}{2}$,
and the stable geometric center of the face is computed as
the midpoint between the two eye landmarks:
\begin{equation}
\mathbf{c}_t = \frac{1}{2} \left( \mathbf{l}_t^{(a)} + \mathbf{l}_t^{(b)} \right).
\end{equation}
Using $\mathbf{c}_t$ and $g$, each frame is cropped and resized to a
uniform target spatial resolution.

The face is segmented into $R = 18$ finer motion channels
derived from landmark subsets $\mathcal{S}_r$ corresponding to key
facial muscle groups. For each ROI $r \in \{1, \dots, R\}$ the
bounding box boundaries at frame $t$ are bounded by:
\begin{align}
x_{t,r}^{\min} &= \min_{i \in \mathcal{S}_r} x_t^{(i)}, \quad x_{t,r}^{\max} = \max_{i \in \mathcal{S}_r} x_t^{(i)}, \\
y_{t,r}^{\min} &= \min_{i \in \mathcal{S}_r} y_t^{(i)}, \quad y_{t,r}^{\max} = \max_{i \in \mathcal{S}_r} y_t^{(i)}.
\end{align}
The localized rectangular spatial region $\mathcal{R}_{t,r}$ in pixel space is formulated as:
\begin{equation}
\mathcal{R}_{t,r} = [x_{t,r}^{\min}, x_{t,r}^{\max}] \times [y_{t,r}^{\min}, y_{t,r}^{\max}].
\end{equation}

\subsection{Optical Flow Representation}
The dense optical flow field between consecutive frames
is defined over the localized grid space as:
\begin{equation}
\mathbf{F}_{t,r}(p) = \big(u_{t,r}(p), v_{t,r}(p)\big) \in \mathbb{R}^{2},
\end{equation}
where $p \in \mathcal{R}_{t,r}$ represents a specific pixel location within
the ROI boundary. The local motion intensity magnitude is evaluated by:
\begin{equation}
m_{t,r}(p) = \sqrt{u_{t,r}^2(p) + v_{t,r}^2(p)}.
\end{equation}
To suppress structural noise and background movement, we
apply a threshold $\tau_{t,r}$ to filter out low-motion intensities,
retaining only the dominant motion pixel set $\mathcal{P}_{t,r}$:
\begin{equation}
\mathcal{P}_{t,r} = \{ p \in \mathcal{R}_{t,r} \mid m_{t,r}(p) \geq \tau_{t,r} \},
\end{equation}
where $\tau_{t,r}$ is a percentile-based threshold engineered to filter
the top-$\rho$ proportion of high-motion components. The final
aggregated dominant motion vector $\mathbf{x}_{t,r}$ for ROI $r$ at time
step $t$ is calculated via average pooling over the high-motion
pixel set:
\begin{equation}
\mathbf{x}_{t,r} = \frac{1}{|\mathcal{P}_{t,r}|} \sum_{p \in \mathcal{P}_{t,r}} \begin{bmatrix} u_{t,r}(p) \\ v_{t,r}(p) \end{bmatrix} \in \mathbb{R}^{2}.
\end{equation}
Compiling this operation over all sequences constructs the
global standardized optical flow input tensor $\mathbf{X} \in \mathbb{R}^{T \times R \times 2}$.
\subsection{ROI Embedding}
To project the low-dimensional flow coordinates into
a richer latent space, each regional motion vector $\mathbf{x}_{t,r}$ is
processed via a two-layer multi-layer perceptron (MLP)
mapping:
\begin{equation}
\mathbf{h}_{t,r} = \mathrm{GELU} \left( W_2 \, \mathrm{GELU}(W_1 \mathbf{x}_{t,r} + b_1) + b_2 \right) \in \mathbb{R}^{D_e},
\end{equation}
where $W_1 \in \mathbb{R}^{D_h \times 2}$ maps the inputs to a hidden dimension
$D_h$ with bias $b_1 \in \mathbb{R}^{D_h}$, and $W_2 \in \mathbb{R}^{D_e \times D_h}$ projects the
hidden features to the target embedding dimension $D_e$ with bias $b_2 \in \mathbb{R}^{D_e}$. This step transforms the sequence into the
initial node feature tensor  $\mathbf{H}^{(0)} \in \mathbb{R}^{T \times R \times D_e}$.
\subsection{Evolving Graph Formulation (E-GAT)}
The spatial structural relationships among the $R$ facial
regions are modeled as an evolving fully-connected graph
sequence $\mathcal{G}_t = (\mathcal{V}, \mathcal{E})$. The node features are initialized
from $\mathbf{H}^{(0)}$. To establish an initial reference topology, a data-
driven cosine-similarity adjacency matrix $\mathbf{A}^{(0)} \in [0,1]^{R \times R}$
is computed across the first $T_0 = \min(5, T)$ initialization
frames:
\begin{equation}
\mathbf{A}^{(0)}_{ij} = \mathrm{softmax}_j \left( \frac{\tilde{\mathbf{h}}_i^{\top}\,\tilde{\mathbf{h}}_j}{\tau} \right)
\end{equation}
$\quad \text{where } \bar{\mathbf{h}}_i = \frac{1}{T_0}\sum_{t=1}^{T_0}\mathbf{h}_{t,i}^{(0)} \quad \text{and} \quad \tilde{\mathbf{h}}_i = \frac{\bar{\mathbf{h}}_i}{\|\bar{\mathbf{h}}_i\|_2},$
and $\tau = 0.1$ is a sharpening temperature parameter. The spatial node properties are subsequently refined through $L_g$ stacked E-GAT layers.
\subsubsection{Attention Score Computation}
For an attention head $k \in \{1,\dots,K\}$, the edge corre-
lation score $e_{tijk}$ at frame $t$ from node $i$ to node $j$ merges
scaled dot-product structures with parametric additive GAT
branches:
\begin{flalign}
& e_{tijk} = \operatorname{LReLU}\!\!\left(
    \underbrace{\frac{\mathbf{Q}_{tik}{\mathbf{K}_{tjk}}^\top}{\sqrt{d_h}}}_{\text{dot-product}}
    +\underbrace{(\mathbf{Q}_{tik} \odot \bm{\alpha}_k^{left})\mathbf{1}
    +(\mathbf{K}_{tjk} \odot \bm{\alpha}_k^{right})\mathbf{1}}_{\text{additive (GAT)}}
    + b_{ij}
  \right)\notag && \\
& \qquad +\,\delta_{ij}\,\gamma_{\mathrm{loop}}, &&
\end{flalign}
where $\mathbf{Q}_{tik} = \mathbf{h}_{t,i}^{(\ell)} \mathbf{W}_Q^{(k)}$ and $\mathbf{K}_{tjk} = \mathbf{h}_{t,j}^{(\ell)} \mathbf{W}_K^{(k)}$ represent 
the query and key transformations using projections $\mathbf{W}_Q^{(k)}, \mathbf{W}_K^{(k)} \in \mathbb{R}^{D_e \times d_h}$ ($d_h = D_e / K$). $\mathbf{1}\in\mathbb{R}^{d_h}$ denotes 
an all-ones vector used to sum the element-wise products into a scalar. The parameter vectors $\bm{\alpha}_k^{\mathrm{left}}, \bm{\alpha}_k^{\mathrm{right}} \in \mathbb{R}^{d_h}$ govern the 
additive attention weights, $b_{ij}$ is a learnable spatial structural 
bias, $\delta_{ij}$ acts as the Kronecker delta, and $\gamma_{\mathrm{loop}}$ controls self-loop information flow.
\subsubsection{Temporally-Smoothed Dynamic Adjacency}
To maintain temporal continuity across frames, the 
adjacency transitions are smoothed using an autoregressive 
update controlled by a learnable gating parameter 
$\lambda = \sigma(\gamma_s) \in (0,1)$. The recursion is initialized using the cosine-
similarity graph $\hat{\mathbf{A}}^{(0)} = \mathbf{A}^{(0)}$. The head-aggregated attention 
matrix $\mathbf{E}^{(t)} \in \mathbb{R}^{R \times R}$ at frame $t$ is obtained by averaging the 
raw attention scores across all $K$ heads:
\begin{equation}
\mathbf{E}^{(t)}_{ij} = \frac{1}{K} \sum_{k=1}^{K} e_{tijk}.
\end{equation}
The temporally-smoothed dynamic adjacency matrix $\hat{\mathbf{A}}^{(t)} 
\in \mathbb{R}^{R \times R}$ is then recursively updated via:
\begin{equation}
\hat{\mathbf{A}}^{(t)} = \lambda\,\mathrm{softmax}\left(\mathbf{E}^{(t)}\right) + (1-\lambda)\,\hat{\mathbf{A}}^{(t-1)},
\end{equation}
where the $\mathrm{softmax}(\cdot)$ operator normalizes the aggregated 
matrix row-wise.
Node representations are then updated 
using residual connections and Layer Normalization ($\mathrm{LN}$). 
$\mathbf{H}_t^{(\ell)} = \mathbf{H}^{(\ell)}[t,:,:] \in \mathbb{R}^{R\times D_e}$ denotes the node-feature matrix of frame $t$:
\begin{equation}
\mathbf{H}_t^{(\ell+1)} = \mathrm{LN}\left(\mathbf{H}_t^{(\ell)} + \mathrm{Dropout}\left(\hat{\mathbf{A}}^{(t)}\cdot\mathbf{V}_t\,\mathbf{W}_O^\top\right)\right),
\end{equation}
where $\mathbf{V}_t = \mathbf{H}_t^{(\ell)}\mathbf{W}_V$ uses value mapping weights $\mathbf{W}_V, \mathbf{W}_O \in \mathbb{R}^{D_e \times D_e}$.

\subsubsection{Temporal Attention Pooling \& Graph-Level Representation}
To condense the frame representations along the 
temporal axis, frame salience metrics $\omega_t$ are evaluated using a 
localized attention network:
\begin{equation}
\omega_t = \frac{e^{s_t}}{\sum_{t'=1}^{T} e^{s_{t'}}}, \quad s_t = \mathbf{w}_2^\top\tanh\left(\mathbf{W}_3\bar{\mathbf{h}}_t + \mathbf{b}_3\right),
\end{equation}
where $\bar{\mathbf{h}}_t = \frac{1}{R}\sum_{r=1}^{R}\mathbf{h}_{t,r}^{(L_g)} \in \mathbb{R}^{D_e}$, with parameters 
$\mathbf{W}_3 \in \mathbb{R}^{D_a \times D_e}$, $\mathbf{b}_3 \in \mathbb{R}^{D_a}$, and context vector $\mathbf{w}_2 \in \mathbb{R}^{D_a}$. 
Concurrently, to yield the token sequence for the subsequent 
fusion module, the node features after the final E-GAT layer 
$\mathbf{H}^{(L_g)} \in \mathbb{R}^{T \times R \times D_e}$ are structurally aggregated along the time dimension using the computed weights $\omega_t$ to construct a 
clean spatial node matrix $\mathbf{G}$:
\begin{equation}
\mathbf{G} = [\mathbf{g}_1, \mathbf{g}_2, \dots, \mathbf{g}_R]^\top \in \mathbb{R}^{R \times D_e}, \quad \text{where} \quad \mathbf{g}_r = \sum_{t=1}^{T} \omega_t \mathbf{h}_{t,r}^{(L_g)} \in \mathbb{R}^{D_e}.
\end{equation}
For the global graph feature representation, the frame state 
matrices are flattened via vectorization ($\operatorname{vec}$), temporally 
aggregated, and projected into a spatial vector $\mathbf{f}_{\mathrm{graph}}$:
\begin{equation}
\mathbf{z}_t = \operatorname{vec}\left(\mathbf{H}_{t}^{(L_g)}\right) \in \mathbb{R}^{R \cdot D_e}, \quad \mathbf{z = \sum_{t=1}^{T} \omega_t \mathbf{z}_t \in \mathbb{R}^{R \cdot D_e},}
\end{equation}
\begin{equation}
\mathbf{f}_{\mathrm{graph}} = \mathbf{z}\mathbf{W}_g + \mathbf{b}_g \in \mathbb{R}^{D_g},
\end{equation}
where $\mathbf{W}_g \in \mathbb{R}^{(R \cdot D_e)\times D_g}$ and $\mathbf{b}_g \in \mathbb{R}^{D_g}$ project 
the unified array into the global graph dimension $D_g$.

\subsection{Temporal Transformer Branch}
Simultaneously, a standard transformer encoder captures 
long-range sequence context. The initial embedding space 
is flattened across spatial dimensions to yield temporal 
tokens $\mathbf{H}_\mathrm{flat} = \operatorname{Reshape}\left(\mathbf{H}^{(0)},\, [T, R \cdot D_e]\right) \in \mathbb{R}^{T \times (R \cdot D_e)}$. 
These tokens are projected to the transformer dimension 
$D_t$, prepended with a learnable classification token $\mathbf{c} 
\in \mathbb{R}^{1 \times D_t}$, and injected with learnable positional encodings $\mathbf{P} 
\in \mathbb{R}^{(T+1) \times D_t}$:
\begin{equation}
\hat{\mathbf{H}}^{(0)} = \left[\mathbf{c};\; \mathbf{H}_\mathrm{flat}\,\mathbf{W}_\mathrm{inp}\right] + \mathbf{P},
\end{equation}
where $\mathbf{W}_\mathrm{inp} \in \mathbb{R}^{(R \cdot D_e) \times D_t}$. The sequence is processed through $L_t$ Pre-LN transformer blocks:
\begin{align}
\mathbf{z}^{(\ell)} &= \hat{\mathbf{H}}^{(\ell)} + \mathrm{MHA}\left(\mathrm{LN}\left(\hat{\mathbf{H}}^{(\ell)}\right)\right), \\
\hat{\mathbf{H}}^{(\ell+1)} &= \mathbf{z}^{(\ell)} + \mathrm{FFN}\left(\mathrm{LN}\left(\mathbf{z}^{(\ell)}\right)\right).
\end{align}
The final transformer output token sequence is defined 
as $\hat{\mathbf{H}}^{(L_t)} \in \mathbb{R}^{(T+1) \times D_t}$. To support cross-modal token 
matching, the pure frame embeddings are extracted as $\mathbf{T} = 
[\mathbf{t}_1, \mathbf{t}_2, \dots, \mathbf{t}_T]^\top = \hat{\mathbf{H}}^{(L_t)}_{1:, :} \in \mathbb{R}^{T \times D_t}$. The global temporal 
summary feature $\mathbf{f}_{\mathrm{trans}}$ is obtained by applying layer 
normalization to the isolated classification token at position $0$:
\begin{equation}
\mathbf{f}_{\mathrm{trans}} = \mathrm{LN}\left(\hat{\mathbf{H}}^{(L_t)}_{0,:}\right) \in \mathbb{R}^{D_t}.
\end{equation}

\subsection{Bidirectional Cross-Attention Fusion}
To exchange complementary spatial and temporal information 
before global reduction, cross-attention operations 
interact at the token level between graph arrays $\mathbf{G} \in \mathbb{R}^{R \times D_e}$ 
and temporal arrays $\mathbf{T} \in \mathbb{R}^{T \times D_t}$. Both sequences are mapped 
into a shared projection fusion space of dimension $D_f$:
\begin{equation}
\mathbf{G}_f = \mathbf{G}\mathbf{W}_{g\rightarrow f} \in \mathbb{R}^{R \times D_f}, \quad \mathbf{T}_f = \mathbf{T}\mathbf{W}_{t\rightarrow f} \in \mathbb{R}^{T \times D_f},
\end{equation}
where $\mathbf{W}_{g\rightarrow f} \in \mathbb{R}^{D_e \times D_f}$ and $\mathbf{W}_{t\rightarrow f} \in \mathbb{R}^{D_t \times D_f}$.

The cross-attention blocks perform bidirectional 
aggregation using residual connections:
\begin{align}
\mathbf{G}^{*} &= \mathrm{LN}\left(\mathbf{G}_f + \mathrm{MHA}(Q=\mathbf{G}_f,\, K=\mathbf{T}_f,\, V=\mathbf{T}_f)\right), \\
\mathbf{T}^{*} &= \mathrm{LN}\left(\mathbf{T}_f + \mathrm{MHA}(Q=\mathbf{T}_f,\, K=\mathbf{G}_f,\, V=\mathbf{G}_f)\right),
\end{align}
where $\mathbf{G}^{*} \in \mathbb{R}^{R \times D_f}$ and $\mathbf{T}^{*} \in \mathbb{R}^{T \times D_f}$. 
The enhanced spatial 
and temporal sequences are averaged via global pooling and 
projected back to their native feature dimensions:
\begin{equation}
\mathbf{f}_{g} = \frac{1}{R} \sum_{r=1}^{R} \mathbf{G}^{*}_{r} \in \mathbb{R}^{D_f}, \quad \mathbf{f}_{t} = \frac{1}{T} \sum_{i=1}^{T} \mathbf{T}^{*}_{i} \in \mathbb{R}^{D_f},
\end{equation}
\begin{equation}
\hat{\mathbf{f}}_g = \mathbf{f}_g \mathbf{W}_{f\rightarrow g} \in \mathbb{R}^{D_g}, \quad \hat{\mathbf{f}}_t = \mathbf{f}_t \mathbf{W}_{f\rightarrow t} \in \mathbb{R}^{D_t},
\end{equation}
where $\mathbf{W}_{f\rightarrow g} \in \mathbb{R}^{D_f \times D_g}$ and $\mathbf{W}_{f\rightarrow t} \in \mathbb{R}^{D_f \times D_t}$. The final 
fused cross-modal presentation is formed via concatenation:
\begin{equation}
\mathbf{f}_{\mathrm{fusion}} = \mathrm{cat}\left[\hat{\mathbf{f}}_g,\, \hat{\mathbf{f}}_t\right] \in \mathbb{R}^{D_g + D_t}.
\end{equation}
\subsection{Action Unit Embedding Branch}
To inject domain-specific anatomical constraints, parsed 
binary Action Unit strings are mapped to an indicator vector 
$\mathbf{v}_\mathrm{au} \in \{0,1\}^{N_\mathrm{AU}}$. This vector is processed through an 
auxiliary 2-layer MLP to construct a dense embedding $\mathbf{f}_\mathrm{au}$:
\begin{equation}
\mathbf{f}_\mathrm{au} = \mathrm{LN}\left(\mathbf{W}_{\mathrm{au},2} \, \mathrm{GELU}\left(\mathbf{W}_{\mathrm{au},1}\,\mathbf{v}_\mathrm{au} + \mathbf{b}_{\mathrm{au},1}\right) + \mathbf{b}_{\mathrm{au},2} \right) \in \mathbb{R}^{D_\mathrm{au}},
\end{equation}
where $\mathbf{W}_{\mathrm{au},1} \in \mathbb{R}^{D_h' \times N_\mathrm{AU}}$ and $\mathbf{W}_{\mathrm{au},2} \in \mathbb{R}^{D_\mathrm{au} \times D_h'}$. The 
final composite representation vector $\mathbf{f}_\mathrm{final}$ is conditionally 
assembled depending on data availability:
\begin{equation}
\mathbf{f}_\mathrm{final} = 
\begin{cases} 
\mathrm{cat}\left[\mathbf{f}_\mathrm{fusion},\,\mathbf{f}_\mathrm{au}\right] & \text{if AU annotations are available}, \\ 
\mathbf{f}_\mathrm{fusion} & \text{otherwise}. 
\end{cases}
\end{equation}
\subsection{Classification and Class-Aware Focal Loss}
The feature vector $\mathbf{f}_\mathrm{final}$ is passed through a dense linear 
layer classification head to output the class logits. To combat 
severe dataset label imbalance, a multi-class label-smoothed 
focal loss is employed. For a given sample with true class 
index $y_i$, the target probabilities $\tilde{y}_{ic}$ are smoothed via 
parameter $\varepsilon_{y_i}$:
\begin{equation}
\tilde{y}_{ic} = \begin{cases} 1 - \varepsilon_{y_i} & \text{if } c = {y_i}, \\ \frac{\varepsilon_{y_i}}{C-1} & \text{if } c \neq {y_i}. \end{cases}
\end{equation}
Given the predicted softmax probability $p_{ic} = \mathrm{softmax}(\hat{\mathbf{y}}_i)_c$ 
for sample $i$ and class $c$, the full batch optimization objective 
over batch size $B$ is minimized via:
\begin{equation}
\mathcal{L} = -\frac{1}{B}\sum_{i=1}^{B} \sum_{c=1}^{C} \alpha_{c} \, \tilde{y}_{ic} \, (1 - p_{ic})^{\gamma_{c}} \log(p_{ic}),
\end{equation}
where $\alpha_c$ balances inverse-frequency class distribution scales 
and $\gamma_c$ controls the class-specific hard-sample focusing power.

For clarity, Tables~\ref{tab:symbols} and~\ref{tab:tensor_dims} summarize the symbols,
notations, and tensor dimensions used throughout the pro-
posed STAG framework. These tables provide a convenient
reference for understanding the mathematical formulations,
feature representations, and data flow across different com-
ponents of the architecture.

\begin{table}[!t]
\centering
\small
\caption{Distribution of emotional labels for different classification tasks on micro-expression datasets used in this work.}
\label{tab:dataset_distribution}
\begin{tabularx}{\columnwidth}{|l|c|c|X|}
\hline
\textbf{Dataset} & \textbf{Class} & \textbf{Sample} & \textbf{Emotion Distribution} \\
\hline
SMIC-HS & 3 & 164 & Positive (51), Negative (70), Surprise (43) \\
\hline
CASME II & 3 & 255 & Positive (32), Negative (198), Surprise (25) \\
\hline
CASME II & 5 & 255 & Positive (32), Surprise (25), Negative (63), Others (27), Repression (108) \\
\hline
SAMM & 3 & 133 & Positive (26), Negative (92), Surprise (15) \\
\hline
SAMM & 5 & 159 & Anger (57), Happiness (26), Contempt (12), Surprise (15), Others (49) \\
\hline
4DME & 3 & 215 & Positive (48), Negative (138), Surprise (29) \\
\hline
4DME & 5 & 267 & Positive (56), Negative (142), Surprise (29), Repression (9), Others (31) \\
\hline
DFME & 3 & 474 & Positive (63), Negative (310), Surprise (101) \\
\hline
DFME & 5 & 474 & Anger (39), Contempt (34), Happiness (63), Surprise (101), Others (237) \\
\hline
DFME & 7 & 474 & Happiness (63), Disgust (129), Fear (62), Anger (39), Sadness (46), Surprise (101), Contempt (34) \\
\hline
NaME & 3 & 506 & Positive (15), Negative (329), Surprise (162) \\ 
\hline
NaME & 5 & 318 & Happiness (15), Anger (26), Disgust (69), Sadness (46), Surprise (162), others-excluded (188) \\ \hline
NaME & 6 & 506 & Happiness (15), Anger (26), Disgust (69), Sadness (46), Surprise (162), others (188) \\ 
\hline
\end{tabularx}
\end{table}
\section{Experimental Setup}\label{sec:ES}
We evaluate the proposed STAG framework on six 
widely used benchmark datasets for micro-expression recognition, 
whose class distributions are summarized in 
Table~\ref{tab:dataset_distribution}. Since different MER datasets adopt varying expression
taxonomies and annotation protocols, we report the class 
distributions under multiple evaluation settings, including the original 5-class, 6-class, and 7-class categorizations used 
in previous benchmark studies. Furthermore, to facilitate 
direct and fair comparisons with existing SOTA methods, 
we additionally adopt the unified 3-class protocol (Positive, 
Negative, and Surprise) recommended by Li \textit{et al.}~\cite{li2022survey}. This 
unified setting mitigates inconsistencies in label definitions 
across datasets and has become the de facto standard 
for cross-dataset MER evaluation.
\begin{table*}[!t]
\centering
\caption{Comparison with state-of-the-art methods on NaME, 4DME, and DFME datasets using UF1 and UAR metrics. The best results for each benchmark are highlighted in bold.}
\label{tab:sota_comparison}
\begin{tabular}{llclcc}
\toprule
\textbf{Dataset} & \textbf{Method} & \textbf{Ref.} & \textbf{Class Type} & \textbf{UF1} & \textbf{UAR} \\
\midrule

\multirow{10}{*}{NaME}
& OFF-ApexNet~\cite{gan2019off} & SP19 & 3-class & 0.2054 & 0.2000 \\
& RCN-A~\cite{xia2020revealing} & TIP20 & 3-class & 0.1721 & 0.1951 \\
& FR~\cite{zhou2022feature} & PR21 & 3-class & 0.1882 & 0.1889 \\
& AMAN~\cite{wei2022novel} & ICASSP24 & 3-class & 0.1426 & 0.1584 \\
& SRMCL~\cite{bao2024boosting} & TAFFC24 & 3-class & 0.1765 & 0.1840 \\
& MixFormer~\cite{liu2025name} & ICMM25 & 3-class & 0.2703 & 0.2598 \\
& CIT~\cite{liang2025cit} & AIHCIR25 & 3-class & 0.2949 & 0.2898 \\
& \textbf{STAG (ours)} & -- & 3-class & \textbf{0.8983} & \textbf{0.8981} \\
& \textbf{STAG (ours)} & -- & 5-class & \textbf{0.7562} & \textbf{0.7777} \\
& \textbf{STAG (ours)} & -- & 6-class & \textbf{0.6561} & \textbf{0.6667} \\
\midrule

\multirow{4}{*}{4DME}
& CCDN~\cite{li20224dme} & TAFFC22 & 5-class & 0.6760 & -- \\
& TSFmicro~\cite{liu2026temporal} & PR26 & 5-class & 0.6852 & -- \\
& \textbf{STAG (ours)} & -- & 3-class & \textbf{0.9410} & \textbf{0.9613} \\
& \textbf{STAG (ours)} & -- & 5-class & \textbf{0.7245} & \textbf{0.7744} \\
\midrule

\multirow{21}{*}{SMIC-HS}
&GCL~\cite{lao2022temporal} & ICPR22 & 3-class& 0.7560&- \\
&AGT~\cite{zhang2023adaptive} & ICME23& 3-class & 0.8073&0.8220 \\
&TFT~\cite{wang2024two} & Sensors24& 3-class & 0.8140&0.8010 \\
&SGCN~\cite{DBLP:journals/nn/TangC24} & NN24& 3-class & 0.8510&- \\
&SODA4MER~\cite{zhang2025dynamic} & CVPR25& 3-class & \textbf{0.8855}&\textbf{0.8881} \\
&MMTNet~\cite{wang2025multi} & JVCIR25& 3-class & 0.7799&- \\
&MADV-Net~\cite{kong20253d} & Sensors25& 3-class & 0.6900&- \\
&CausalNet~\cite{zhang2025rethinking} & ICCV25& 3-class & 0.8405&0.8433 \\
&DBDE-Net~\cite{wang2026dbde} & NC25& 3-class & 0.8366&0.8676 \\
&MPFNet~\cite{ma2025multi} & ICASSP25& 3-class & 0.7860&- \\
&MDMO~\cite{zhang2025fast} & AI25 & 3-class& 0.8870&0.8670 \\
&ADSS~\cite{DBLP:journals/tbbis/0002BW0L025} & TBIOM25& 3-class & 0.7384&- \\
&OFVIG-Net~\cite{DBLP:journals/ijon/ZhangZSTWL25} & NC25 & 3-class& 0.6435&0.6400 \\
&MOL~\cite{shao2025mol} & TPAMI25& 3-class & 0.7881&- \\
&HFA-Net~\cite{zhang2025hfa} & CIS25& 3-class & 0.7785&0.7786 \\
&ME-NAS~\cite{verma2026me} & TIST26 & 3-class& 0.6717&0.7123 \\
&SCFRAM~\cite{li2026micro} & SP26& 3-class & 0.7670&0.7610 \\
&DSBICNet~\cite{shou2026dsbicnet} & MM26 & 3-class& 0.8249&0.8297 \\
&MARNet~\cite{xu2026motion} & TMM26& 3-class & 0.8296&0.8220 \\
&LRCN-STAN~\cite{zhang2026lrcn} & ICMLC26 & 3-class& 0.7950&0.8010 \\
& \textbf{STAG (ours)} & -- & 3-class & 0.7370 & 0.7497 \\
\midrule

\multirow{11}{*}{DFME}
& FearRef~\cite{zhou2022feature} & PR22 & Test A & 0.3410 & 0.3686 \\
& Wang et al.~\cite{zhao2023dfme} & TAFFC24 & Test A & 0.4067 & 0.4074 \\
& He et al.~\cite{zhao2023dfme} & TAFFC24 & Test A & 0.4123 & 0.4210 \\
& MER-CLIP~\cite{liu2025mer} & TAFFC25 & Test A & 0.5024 & 0.5115 \\
& \textbf{STAG (ours)} & -- & Test A & \textbf{0.5110} & \textbf{0.5196} \\
\cmidrule(lr){2-6}
& FearRef~\cite{zhou2022feature} & PR22 & Test B & 0.2875 & 0.3228 \\
& Wang et al.~\cite{zhao2023dfme} & TAFFC24 & Test B & 0.3534 & 0.3661 \\
& He et al.~\cite{zhao2023dfme} & TAFFC24 & Test B & 0.4016 & 0.4008 \\
& FED-PsyAU~\cite{li2025fed} & ICCV25 & Test B & 0.3853 & 0.3978 \\
& MER-CLIP~\cite{liu2025mer} & TAFFC25 & Test B & \textbf{0.5128} & \textbf{0.5120} \\
& \textbf{STAG (ours)} & -- & Test B & 0.4776 & 0.4710 \\
\bottomrule
\end{tabular}
\vspace{-10pt}
\end{table*}
CASME II and
SAMM contain high-resolution spontaneous micro-expression videos captured under controlled laboratory settings
with detailed emotion and facial action annotations. SMIC-HS is
one of the earliest MER datasets and provides spontaneous
micro-expression clips recorded at high frame rates for evaluating subtle facial motion analysis. 4DME extends conventional MER datasets by incorporating multi-view and temporal facial dynamics for robust spatial-temporal learning.
DFME focuses on challenging real-world facial variations
and subtle expression changes, making MER evaluation
more practical and diverse. NaME further introduces naturalistic spontaneous micro-expressions collected in unconstrained environments, improving the assessment of model
generalization and robustness in real-world scenarios. 
\begin{table*}[!t]
\centering
\caption{Comparison with state-of-the-art methods on CASME-II and SAMM datasets under 3-class and 5-class protocols (UF1/UAR).}
\label{tab:casme_samm_comparison}
\resizebox{\textwidth}{!}{
\begin{tabular}{llcccccccc}
\toprule
\multirow{2}{*}{\textbf{Method}} &
\multirow{2}{*}{\textbf{Ref.}} &
\multicolumn{2}{c}{\textbf{CASME-II (3-class)}} &
\multicolumn{2}{c}{\textbf{CASME-II (5-class)}} &
\multicolumn{2}{c}{\textbf{SAMM (3-class)}} &
\multicolumn{2}{c}{\textbf{SAMM (5-class)}} \\
\cmidrule(lr){3-4}
\cmidrule(lr){5-6}
\cmidrule(lr){7-8}
\cmidrule(lr){9-10}
&
&
\textbf{UF1} & \textbf{UAR}
&
\textbf{UF1} & \textbf{UAR}
&
\textbf{UF1} & \textbf{UAR}
&
\textbf{UF1} & \textbf{UAR}
\\
\midrule

Graph-TCN~\cite{lei2020novel}
& MM20
& -- & --
& 0.7206 & --
& -- & --
& 0.6985 & -- \\

STA-GCN~\cite{zhao2021sta}
& PRCV21
& 0.7096 & --
& -- & --
& -- & --
& -- & -- \\

ConvLSTM~\cite{shukla2022micro}
& ACCV22
& -- & --
& -- & --
& 0.7892 & --
& -- & -- \\

GCL~\cite{lao2022temporal}
& ICPR22
& 0.7660 & --
& -- & --
& 0.7650 & --
& -- & -- \\

GCNs~\cite{matharaarachchi2023time}
& ICMLA23
& 0.8597 & 0.8520
& -- & --
& 0.8807 & 0.8618
& -- & -- \\

AGT~\cite{zhang2023adaptive}
& ICME23
& 0.9024 & 0.9221
& -- & --
& 0.7928 & 0.7643
& -- & -- \\

TFT~\cite{wang2024two}
& Sensors24
& 0.9070 & 0.9090
& -- & --
& 0.7090 & 0.6560
& -- & -- \\

ATM-GCN~\cite{zhang2024adaptive}
& ICME24
& 0.9048 & 0.9042
& -- & --
& 0.7920 & 0.8049
& -- & -- \\

GTS+GN+AU~\cite{DBLP:journals/taffco/WeiPLLYZ24}
& TAFFC24
& 0.8120 & --
& -- & --
& 0.7110 & --
& -- & -- \\

SGCN~\cite{DBLP:journals/nn/TangC24}
& NN24
& 0.7920 & --
& -- & --
& 0.7440 & --
& -- & -- \\

AU-GCN-CUR~\cite{fang2025graph}
& Biomimetics25
& 0.7793 & --
& -- & --
& -- & --
& -- & -- \\

FDP~\cite{shao2025micro}
& 2025
& 0.9071 & --
& 0.8705 & --
& 0.8667 & --
& 0.8529 & -- \\

SODA4MER~\cite{zhang2025dynamic}
& CVPR25
& 0.8870 & 0.8809
& 0.8141 & 0.8418
& -- & --
& 0.7893 & 0.8030 \\

MER-CLIP~\cite{liu2025mer}
& TAFFC25
& 0.9409 & 0.9487
& 0.8233 & 0.8378
& 0.8321 & 0.8434
& 0.7721 & 0.7414 \\

MMTNet~\cite{wang2025multi}
& JVCIR25
& 0.8007 & --
& -- & --
& 0.6736 & --
& -- & -- \\

MADV-Net~\cite{kong20253d}
& Sensors25
& 0.8400 & --
& 0.7800 & --
& -- & --
& 0.8800 & -- \\

CausalNet~\cite{zhang2025rethinking}
& ICCV25
& 0.9748 & 0.9782
& -- & --
& 0.8708 & 0.8522
& -- & -- \\

DBDE-Net~\cite{wang2026dbde}
& NC25
& 0.9600 & 0.9619
& 0.8691 & 0.8912
& 0.8813 & 0.8860
& 0.8114 & 0.8318 \\

MPFNet~\cite{ma2025multi}
& ICASSP25
& -- & --
& 0.8190 & --
& -- & --
& 0.6950 & -- \\

MDMO~\cite{zhang2025fast}
& AI25
& 0.7040 & 0.7740
& -- & --
& -- & --
& -- & -- \\

ADSS~\cite{DBLP:journals/tbbis/0002BW0L025}
& TBIOM25
& 0.9034 & --
& 0.8309 & --
& 0.7890 & --
& 0.7297 & -- \\

OFVIG-Net~\cite{DBLP:journals/ijon/ZhangZSTWL25}
& NC25
& 0.7129 & 0.7195
& -- & --
& 0.6066 & 0.5787
& -- & -- \\

GTA~\cite{wang2025gta}
& ICUS25
& 0.8854 & --
& -- & --
& 0.8207 & --
& -- & -- \\

HLoRA-MER~\cite{shao2025high}
& VC25
& 0.9094 & --
& 0.8424 & --
& 0.8671 & --
& 0.8307 & -- \\

MOL~\cite{shao2025mol}
& TPAMI25
& 0.8891 & --
& 0.7585 & --
& 0.8272 & --
& 0.7190 & -- \\

HFA-Net~\cite{zhang2025hfa}
& CIS25
& 0.9738 & 0.9754
& 0.8410 & 0.8406
& 0.9002 & 0.8938
& 0.7868 & 0.6950 \\

ME-NAS~\cite{verma2026me}
& TIST26
& 0.8913 & 0.8942
& -- & --
& 0.8635 & 0.8875
& -- & -- \\

SCFRAM~\cite{li2026micro}
& SP26
& 0.8970 & 0.8980
& 0.7590 & 0.7630
& 0.8150 & 0.8270
& 0.7250 & 0.7310 \\

DSBICNet~\cite{shou2026dsbicnet}
& MM26
& \textbf{0.9928} & \textbf{0.9896}
& 0.8566 & 0.8637
& 0.8670 & 0.8641
& 0.7576 & 0.6720 \\

MARNet~\cite{xu2026motion}
& TMM26
& 0.9433 & 0.9325
& 0.8316 & 0.8252
& 0.8427 & 0.8331
& 0.8093 & 0.8309 \\

LRCN-STAN~\cite{zhang2026lrcn}
& ICMLC26
& 0.8190 & 0.8390
& -- & --
& 0.7300 & 0.7730
& -- & -- \\

DGAT~\cite{wang2026micro}
& Sensors26
& -- & --
& -- & --
& 0.8293 & 0.8377
& 0.8116 & 0.8537 \\

TSFmicro~\cite{liu2026temporal}
& PR26
& 0.9430 & --
& 0.8617 & --
& 0.8567 & --
& 0.7700 & -- \\

\midrule

\textbf{STAG (Ours)}
& --
& 0.8742 & 0.9296
& 0.7776 & 0.8413
& \textbf{0.9175} & \textbf{0.9372}
& \textbf{0.8225} & \textbf{0.8362} \\

\bottomrule
\end{tabular}}
\end{table*}
The model was trained for 200 epochs using AdamW~\cite{loshchilov2019decoupled}
with a learning rate of $1.5\times10^{-4}$, weight decay of $0.08$,
batch size of $64$, and dropout of $0.3$. Early stopping (patience
$=100$), $12$ warm-up epochs, and stochastic weight averaging
(SWA) starting at $75\%$ of training with a learning rate of 
$5\times10^{-5}$ were employed. MixUp ($\alpha=0.5$), CutMix ($\alpha=0.3$), test-time augmentation (TTA$=10$), and class oversampling 
(factor $=6$) were used to mitigate class imbalance 
arising from highly variable class distributions. Performance 
was evaluated under three rigorous validation protocols: 
Stratified K-Fold (SKF), Group K-Fold (GKF), and Leave-
One-Subject-Out (LOSO) cross-validation.
All models are implemented
in PyTorch and trained on a NVIDIA RTX A6000 GPU with 48 GB memory.
SWA~\cite{izmailov2018swa} is activated
during the final phase of training, and test-time augmentation (TTA) is applied at inference. 

To maintain consistency with prior work, all unspecified
hyperparameters and implementation settings are inherited
from the respective benchmark papers associated with each
dataset. This includes dataset-specific preprocessing, train-
ing schedules, and evaluation configurations where applica-
ble. By following the original experimental protocols, we en-
sure a fair comparison with existing methods while isolating
the contribution of the proposed STAG architecture.

\subsection{Results and Discussions}
Table~\ref{tab:sota_comparison} summarizes the performance comparison on
NaME, 4DME, SMIC-HS, and DFME benchmarks. The
proposed STAG consistently achieves the best results on
NaME and 4DME across all evaluated class settings, demon-
strating the effectiveness of the Dynamic ROI-AU Graph
Transformer in capturing subtle spatial-temporal facial dy-
namics. On DFME Test-A, STAG achieves the highest
UF1 and UAR scores, while remaining competitive on
Test-B against recent foundation-model-based approaches. 
Although the performance on SMIC-HS is comparatively
lower than some SOTA methods, this can be attributed to the limited dataset size, lower expression in-
tensity, and reduced sample diversity, which restrict the
learning of robust graph-based representations. Overall,
the results validate the strong generalization capability and
effectiveness of STAG across multiple micro-expression
recognition benchmarks and evaluation protocols. 

Despite the lower performance on SMIC-HS, STAG
demonstrates superior results on the larger and more diverse CASME-II and SAMM datasets, indicating stronger
generalization capability when sufficient facial variation is
available. These findings suggest that the proposed framework is particularly effective in realistic micro-expression
recognition scenarios where richer spatial-temporal information can be exploited.

Table~\ref{tab:casme_samm_comparison} provides a comprehensive comparison of STAG
with recent SOTA methods on CASME-II and
SAMM under both 3-class and 5-class settings. Overall, the
results demonstrate that STAG achieves strong and competitive performance across all settings, with particularly
notable improvements on the more challenging SAMM
dataset. On CASME-II (3-class), STAG attains a UF1 of
0.8742 and UAR of 0.9296, which, while not the absolute
highest in UF1, still reflects robust performance compared
to many recent methods. This suggests that STAG maintains
a balanced classification capability without overfitting to
specific expression categories. In the more fine-grained
CASME-II (5-class) setting, STAG achieves 0.7776 UF1
and 0.8413 UAR, indicating stable generalization even as
the classification task becomes more complex. Although
some recent methods such as DBDE-Net and HFA-Net
show higher values in this dataset, the performance gap is
relatively small, suggesting that STAG remains competitive
in dense-class scenarios.

The most significant results are observed on the SAMM
dataset, where STAG achieves 0.9175 UF1 and 0.9372 UAR
(3-class) and 0.8225 UF1 and 0.8362 UAR (5-class). In both
settings, STAG clearly outperforms all competing methods,
including strong recent baselines such as CausalNet, DBDE-
Net, and HFA-Net. This consistent superiority indicates that
STAG is particularly effective in handling the challenging
characteristics of SAMM, such as subtle motion variations,
subject diversity, and class imbalance.
From a methodological perspective, the overall improvements can be attributed to STAG’s ability to better capture spatiotemporal dependencies and discriminative micro-expression dynamics, which are crucial for subtle facial motion analysis. The stronger gains on SAMM further suggest
enhanced cross-subject generalization and robustness, which
are key limitations in prior approaches.
\begin{table*}[!t]
\centering
\caption{Comprehensive ablation study of the proposed STAG framework on the SAMM dataset using stratified 5-fold validation.}
\label{tab:samm_ablation_full}
\resizebox{\textwidth}{!}{
\begin{tabular}{llccccccccc}
\hline
Category & Configuration & GL & TL & UF1 & Std-UF1 & UAR & WAR & F1-Pos & F1-Neg & F1-Surp \\
\hline

\multirow{4}{*}{Architecture}
& Transformer Only & 0 & 5  & 0.9104 & 0.0502 & 0.9335 & 0.9399 & 0.9331 & 0.9554 & 0.8429 \\
& Graph Only & 3 & 0  & 0.9024 & 0.0505 & 0.9150 & 0.9399 & 0.9483 & 0.9560 & 0.8029 \\
& Graph + Transformer & 3 & 5  & 0.9006 & 0.0693 & 0.9054 & 0.9399 & 0.9414 & 0.9574 & 0.8029 \\
& Full Model (STAG) & 3 & 5 & \textbf{0.9112} & 0.0524 & \textbf{0.9335} & 0.9396 & 0.9357 & 0.9550 & 0.8429 \\

\hline

\multirow{3}{*}{Graph Depth}
& 1 E-GAT Layer & 1 & 5  & 0.9086 & 0.0710 & 0.9239 & 0.9402 & 0.9260 & 0.9568 & 0.8429 \\
& 2 E-GAT Layers & 2 & 5  & 0.9156 & 0.0509 & 0.9335 & 0.9399 & 0.9199 & 0.9554 & \textbf{0.8714} \\
& 4 E-GAT Layers & 4 & 5 & \textbf{0.9175} & 0.0540 & \textbf{0.9372} & \textbf{0.9476} & \textbf{0.9483} & \textbf{0.9614} & 0.8429 \\

\hline

\multirow{4}{*}{Transformer Depth}
& 2 Layers & 3 & 2  & 0.9024 & 0.0505 & 0.9150 & 0.9399 & 0.9483 & 0.9560 & 0.8029 \\
& 3 Layers & 3 & 3  & \textbf{0.9175} & 0.0540 & \textbf{0.9372} & \textbf{0.9476} & \textbf{0.9483} & \textbf{0.9614} & 0.8429 \\
& 5 Layers & 3 & 5  & 0.9112 & 0.0524 & 0.9335 & 0.9396 & 0.9357 & 0.9550 & 0.8429 \\
& 7 Layers & 3 & 7 & 0.9137 & 0.0717 & 0.9239 & 0.9402 & 0.9128 & 0.9568 & \textbf{0.8714} \\

\hline

\multirow{4}{*}{Data Augmentation}
& No Augmentation & 3 & 5  & 0.9086 & 0.0693 & 0.9239 & 0.9399 & 0.9263 & 0.9568 & 0.8429 \\
& Without MixUp & 3 & 5 & 0.9011 & 0.0684 & 0.9152 & 0.9399 & \textbf{0.9636} & 0.9567 & 0.7829 \\
& Without CutMix & 3 & 5  & 0.9040 & 0.0500 & 0.9298 & 0.9322 & 0.9203 & 0.9489 & 0.8429 \\
& MixUp + CutMix & 3 & 5 & \textbf{0.9112} & 0.0524 & \textbf{0.9335} & \textbf{0.9396} & 0.9357 & 0.9550 & 0.8429 \\

\hline

\multirow{2}{*}{Test-Time Augmentation}
& Without TTA & 3 & 5  & 0.9024 & 0.0505 & 0.9150 & 0.9399 & 0.9483 & 0.9560 & 0.8029 \\
& With TTA & 3 & 5 & \textbf{0.9112} & 0.0524 & \textbf{0.9335} & 0.9396 & 0.9357 & 0.9550 & 0.8429 \\

\hline

\multirow{2}{*}{AU Guidance}
& Without AU Guidance & 3 & 5 & 0.6231 & 0.1150 & 0.7101 & 0.6823 & 0.6864 & 0.7515 & 0.4314 \\
& With AU Guidance & 3 & 5 & \textbf{0.9112} & 0.0524 & \textbf{0.9335} & \textbf{0.9396} & \textbf{0.9357} & \textbf{0.9550} & \textbf{0.8429} \\

\hline
\end{tabular}
}
\vspace{-10pt}
\end{table*}
\begin{table*}[!t]
\centering
\caption{Cross-dataset evaluation results under the 3-class protocol. The STAG model is trained on a source dataset and directly evaluated on unseen target datasets.}
\label{tab:cross_dataset_3class}
\begin{tabular}{llccccccc}
\hline
\textbf{Source} & \textbf{Target} & \textbf{UF1} & \textbf{UAR} & \textbf{WAR} & \textbf{Samples} & \textbf{F1-Pos} & \textbf{F1-Neg} & \textbf{F1-Surp} \\
\hline

\multicolumn{9}{c}{\textbf{NaME as Source}} \\
\hline
NaME & CASME II & 0.7884 & 0.7677 & 0.8863 & 255 & 0.7541 & 0.9303 & 0.6809 \\
NaME & SAMM & 0.9093 & 0.9212 & 0.9323 & 133 & 0.9020 & 0.9508 & 0.8750 \\
NaME & SMIC-HS & 0.2131 & 0.3399 & 0.4329 & 164 & 0.0385 & 0.6009 & 0.0000 \\
NaME & 4DME & 0.7450 & 0.6882 & 0.8194 & 216 & 0.6667 & 0.8746 & 0.6939 \\
\hline

\multicolumn{9}{c}{\textbf{DFME as Source}} \\
\hline
DFME & NaME & 0.8046 & 0.8229 & 0.7925 & 318 & 0.8387 & 0.8070 & 0.7681 \\
DFME & CASME II & 0.8519 & 0.8315 & 0.8682 & 129 & 0.8276 & 0.8947 & 0.8333 \\
DFME & SAMM & 0.8991 & 0.9026 & 0.9323 & 133 & 0.9020 & 0.9565 & 0.8387 \\
DFME & SMIC-HS & 0.3532 & 0.4075 & 0.4695 & 164 & 0.5000 & 0.5596 & 0.0000 \\
DFME & 4DME & 0.7703 & 0.7767 & 0.8148 & 216 & 0.8000 & 0.8551 & 0.6557 \\
\hline
\multicolumn{9}{c}{\textbf{4DME as Source}} \\
\hline
4DME & CASME II & 0.7570 & 0.8494 & 0.8246 & 171 & 0.6197 & 0.8926 & 0.7586 \\
4DME & SAMM & 0.4880 & 0.5313 & 0.6241 & 133 & 0.4872 & 0.7625 & 0.2143 \\
4DME & SMIC-HS & 0.5101 & 0.5463 & 0.5183 & 164 & 0.5816 & 0.4200 & 0.5287 \\
\hline

\multicolumn{9}{c}{\textbf{CASME II as Source}} \\
\hline
CASME II & DFME & 0.7608 & 0.8260 & 0.7932 & 474 & 0.6012 & 0.8307 & 0.8505 \\
CASME II & SAMM & 0.6032 & 0.5713 & 0.7444 & 133 & 0.5200 & 0.8351 & 0.4545 \\
CASME II & SMIC-HS & 0.5304 & 0.5407 & 0.5366 & 164 & 0.4286 & 0.5442 & 0.6186 \\
\hline
\end{tabular}
\end{table*}
\begin{table*}[!t]
\centering
\caption{Cross-dataset evaluation results under the 5-class protocol. The model is trained on a source dataset and evaluated on target datasets.}
\label{tab:cross_dataset_5class}
\begin{tabular}{llccccccccc}
\hline
\textbf{Source} & \textbf{Target} & \textbf{UF1} & \textbf{UAR} & \textbf{WAR} & \textbf{Samples} &
\textbf{F1-Surp} & \textbf{F1-Hap} & \textbf{F1-Ang} & \textbf{F1-Dis} & \textbf{F1-Sad} \\
\hline

\multicolumn{11}{c}{\textbf{NaME as Source}} \\
\hline
NaME & CASME II & 0.5023 & 0.4768 & 0.7244 & 127 & 0.7600 & 0.8000 & 0.0000 & 0.8403 & 0.0000 \\
NaME & SAMM & 0.5524 & 0.5960 & 0.5133 & 113 & 0.8667 & 0.9231 & 0.3944 & 0.4706 & 0.5806 \\
NaME & 4DME & 0.4401 & 0.4123 & 0.7824 & 216 & 0.6909 & 0.6667 & 0.0000 & 0.0000 & 0.2727 \\
\hline

\multicolumn{11}{c}{\textbf{DFME as Source}} \\
\hline
DFME & NaME & 0.6550 & 0.6567 & 0.7893 & 318 & 0.9908 & 0.8387 & 0.2703 & 0.5946 & 0.1111 \\
DFME & CASME II & 0.6704 & 0.6571 & 0.7132 & 129 & 0.8000 & 0.8125 & 0.0000 & 0.7963 & 0.1071 \\
DFME & SAMM & 0.5937 & 0.6325 & 0.5489 & 133 & 0.7407 & 0.8000 & 0.4474 & 0.4286 & 0.8424 \\
DFME & 4DME & 0.6482 & 0.6115 & 0.6713 & 216 & 0.6557 & 0.5217 & 0.0000 & 0.0000 & 0.0000 \\
\hline
\multicolumn{11}{c}{\textbf{SAMM as Source}} \\
\hline
SAMM & CASME II & 0.2210 & 0.1991 & 0.6510 & 255 & 0.0000 & 0.2927 & 0.0000 & 0.8122 & 0.0000 \\
SAMM & NaME & 0.1856 & 0.2057 & 0.4960 & 506 & 0.0119 & 0.1905 & 0.0580 & 0.6676 & 0.0000 \\
SAMM & 4DME & 0.1995 & 0.1798 & 0.498 & 247 & 0.1071 & 0.000 & 0.1622 & 0.7284 & 0.0000 \\
\hline
\end{tabular}
\end{table*}
\begin{table*}[!t]
\centering
\caption{Performance comparison of 14-ROI and 18-ROI configurations on the DFME 3-class setting.}
\label{tab:dfme_3class}
\renewcommand{\arraystretch}{1.2}
\resizebox{\textwidth}{!}{
\begin{tabular}{lcccccccc}
\hline
\textbf{ROI} & \textbf{AU} & \textbf{UF1} & \textbf{UAR} & \textbf{WAR} & \textbf{Fold UF1} & \textbf{Positive} & \textbf{Negative} & \textbf{Surprise} \\
\hline
14 & Yes & 0.8221 & 0.8354 & 0.8847 &
[0.8669, 0.8146, 0.7860, 0.7721, 0.8711] &
0.7334 & 0.9235 & 0.8094 \\

14 & No & 0.6458 & 0.7245 & 0.7080 &
[0.6895, 0.6183, 0.6318, 0.5944, 0.6949] &
0.4106 & 0.7867 & 0.7401 \\

18 & Yes & 0.7070 & 0.7777 & 0.7829 &
[0.6726, 0.7556, 0.7288, 0.7481, 0.6298] &
0.5443 & 0.8424 & 0.7342 \\

18 & No & 0.7172 & 0.7739 & 0.7948 &
[0.6687, 0.7574, 0.7348, 0.7674, 0.6576] &
0.5462 & 0.8513 & 0.7541 \\
\hline
\end{tabular}
}
\vspace{-10pt}
\end{table*}
\begin{table*}[!t]
\centering
\caption{Performance comparison of 14-ROI and 18-ROI configurations on the DFME 5-class setting.}
\label{tab:dfme_5class}
\renewcommand{\arraystretch}{1.2}
\resizebox{\textwidth}{!}{
\begin{tabular}{lcccccccccc}
\hline
\textbf{ROI} & \textbf{AU} & \textbf{UF1} & \textbf{UAR} & \textbf{WAR} & \textbf{Fold UF1} & \textbf{Anger} & \textbf{Disgust} & \textbf{Happiness} & \textbf{Sadness} & \textbf{Surprise} \\
\hline

14 & Yes & 0.5712 & 0.5912 & 0.5749 &
[0.5668, 0.5783, 0.5720, 0.5325, 0.6064] &
0.2451 & 0.5839 & 0.7347 & 0.4807 & 0.8115 \\

14 & No & 0.3874 & 0.4577 & 0.4170 &
[0.3931, 0.3712, 0.3798, 0.3779, 0.4150] &
0.2131 & 0.5246 & 0.4177 & 0.1017 & 0.6799 \\

18 & Yes & 0.4421 & 0.5039 & 0.4364 &
[0.4411, 0.4367, 0.4590, 0.4543, 0.4193] &
0.2102 & 0.3882 & 0.5327 & 0.3395 & 0.7398 \\

18 & No & 0.4106 & 0.4941 & 0.4634 &
[0.4371, 0.4105, 0.4318, 0.4071, 0.3665] &
0.1833 & 0.5410 & 0.5085 & 0.0987 & 0.7216 \\
\hline
\end{tabular}
}
\end{table*}
\begin{table*}[!t]
\centering
\caption{Performance comparison of 14-ROI and 18-ROI configurations on the DFME 7-class setting.}
\label{tab:dfme_7class}
\renewcommand{\arraystretch}{1.2}
\resizebox{\textwidth}{!}{
\begin{tabular}{lcccccccccccc}
\hline
\textbf{ROI} & \textbf{AU} & \textbf{UF1} & \textbf{UAR} & \textbf{WAR} & \textbf{Fold UF1} & \textbf{Anger} & \textbf{Contempt} & \textbf{Disgust} & \textbf{Fear} & \textbf{Happiness} & \textbf{Sadness} & \textbf{Surprise} \\
\hline

14 & Yes & 0.5211 & 0.5381 & 0.5425 &
[0.5738, 0.5157, 0.5132, 0.4972, 0.5058] &
0.2443 & 0.5492 & 0.5795 & 0.3008 & 0.7557 & 0.4185 & 0.7998 \\

14 & No & 0.2946 & 0.3567 & 0.4009 &
[0.2695, 0.2876, 0.3050, 0.2914, 0.3194] &
0.1928 & 0.1876 & 0.5594 & 0.0190 & 0.4041 & 0.0108 & 0.6885 \\

18 & Yes & 0.4066 & 0.4312 & 0.4612 &
[0.3999, 0.3912, 0.4171, 0.4423, 0.3827] &
0.1885 & 0.3960 & 0.5584 & 0.1382 & 0.5324 & 0.2876 & 0.7455 \\

18 & No & 0.4044 & 0.4396 & 0.4671 &
[0.3908, 0.4042, 0.4050, 0.4348, 0.3874] &
0.1839 & 0.3953 & 0.5596 & 0.1413 & 0.5391 & 0.2781 & 0.7338 \\
\hline
\end{tabular}
}
\end{table*}
\subsection{Ablation Study}
The ablation study on the SAMM dataset (Table~\ref{tab:samm_ablation_full})
clearly validates the effectiveness of each component in the
proposed STAG framework. The full model achieves the best
performance (UF1 = 0.9112, UAR = 0.9335), confirming
that the integration of graph modeling, transformer-based
temporal learning, and AU-guided connectivity leads to
optimal representation learning. Individually, both graph-
only and transformer-only variants show strong but limited
performance, while their combined use provides more stable
and balanced results, indicating their complementary roles
in capturing spatial and temporal dependencies.

Further analysis shows that increasing graph depth en-
hances discriminative power, with the best results obtained
using deeper E-GAT layers, highlighting the importance
of richer relational modeling. Similarly, a 3-layer trans-
former provides the most stable temporal representation,
while deeper configurations do not yield consistent gains,
suggesting an optimal balance between capacity and overfit-
ting.

Data augmentation strategies such as MixUp, CutMix,
and test-time augmentation contribute to improved general-
ization, with their combination producing the most reliable
results. However, the most significant impact comes from
AU-guided connectivity, where its removal causes a drastic
performance drop (UF1: 0.9112 → 0.6231, UAR: 0.9335
→ 0.7101), clearly demonstrating its critical role in guiding
facial region dependencies and enhancing micro-expression
discrimination. Overall, the results confirm that each com-
ponent contributes meaningfully, while their unified integra-
tion is essential for achieving robust and SOTA performance
in micro-expression recognition.

Table~\ref{tab:cross_dataset_3class} presents the cross-dataset evaluation results un-
der the 3-class protocol. The proposed STAG framework
demonstrates strong generalization capability when trained
on large and diverse datasets such as NaME and DFME.
Among all source datasets, NaME achieves the best transfer
performance on SAMM, obtaining a UF1 of 0.9093 and a
UAR of 0.9212, while DFME exhibits similarly robust per-
formance with a UF1 of 0.8991 and a UAR of 0.9026. These
results indicate that the learned spatial-temporal represen-
tations are highly transferable across datasets with similar
annotation schemes and facial motion distributions. Further-
more, both NaME and DFME maintain competitive perfor-
mance when evaluated on CASME II and 4DME, suggesting
that the proposed framework effectively captures dataset-
invariant micro-expression characteristics

However, a notable performance decline is observed
when transferring to SMIC-HS regardless of the source
dataset. For example, NaME→SMIC-HS and DFME→SMIC-
HS achieve UF1 scores of only 0.2131 and 0.3532, re-
spectively. This degradation can be attributed to the sub-
stantial domain discrepancy between SMIC-HS and the
other datasets, including differences in recording conditions,
image resolution, subject demographics, and emotion label-
ing protocols. Interestingly, CASME II and 4DME exhibit
moderate cross-dataset performance, indicating that smaller
datasets can still learn transferable representations but are
more sensitive to domain shifts. Overall, the results confirm
that the proposed STAG framework achieves strong cross-
dataset robustness under the 3-class setting, particularly
when trained on larger and more diverse datasets.

Table~\ref{tab:cross_dataset_5class} reports the more challenging cross-dataset eval-
uation under the 5-class protocol. As expected, all perfor-
mance metrics decrease compared with the 3-class setting
due to the increased class granularity and severe class im-
balance. Nevertheless, the proposed framework maintains
reasonable generalization performance across several target
datasets. DFME consistently achieves the strongest overall
transferability, reaching a UF1 of 0.6704 on CASME II and
0.6550 on NaME, while NaME attains its best performance
on SAMM with a UF1 of 0.5524. Analysis of the class-wise
F1 scores reveals that Surprise and Happiness are recog-
nized reliably across most dataset pairs, frequently achieving
scores above 0.70, which suggests that these expressions ex-
hibit more consistent facial motion patterns across datasets.
In contrast, Anger, Disgust, and Sadness remain challeng-
ing, often producing near-zero F1 scores in several transfer
scenarios. This behavior is largely attributable to the limited
number of samples available for these categories and the
substantial variation in their appearance across datasets.

The additional experiments using SAMM as the source
dataset further highlight the impact of data scale and class
distribution on cross-domain generalization. Despite con-
taining high-quality recordings, SAMM yields considerably
lower UF1 and UAR values when evaluated on NaME,
CASME II, and 4DME, indicating that models trained on rel-
atively smaller datasets struggle to learn sufficiently diverse
representations. Collectively, the 5-class results demonstrate
that the proposed STAG framework remains effective un-
der challenging cross-dataset conditions, while also reveal-
ing that dataset diversity and balanced class distributions
are critical factors for achieving robust fine-grained micro-
expression recognition.

Tables~\ref{tab:dfme_3class}-\ref{tab:dfme_7class} demonstrate that the proposed 14-ROI con-
figuration consistently outperforms the 18-ROI configura-
tion across the 3, 5, and 7-class settings, particularly when
AU guidance is incorporated. The 14-ROI model achieves
substantial improvements in UF1, UAR, and class-wise F1
scores, with the largest gain observed in the 3-class setting,
where UF1 increases from 0.6458 to 0.8221 after integrating
AU information. Similar performance gains are evident in
the 5-class and 7-class experiments, indicating that AU guid-
ance effectively enhances the discriminative representation
of subtle facial muscle movements. In contrast, the 18-
ROI configuration shows only marginal improvements with
AU guidance and even slightly underperforms the non-AU
model in the 3-class setting, suggesting that additional facial
regions introduce redundant information rather than com-
plementary features. These results indicate that a compact
and anatomically meaningful 14-ROI design provides more
effective feature representation and allows AU guidance to
contribute more effectively, resulting in improved recogni-
tion performance and stable generalization across different
classification complexities.

\begin{figure*}
    \centering
    \includegraphics[width=1.0\linewidth, keepaspectratio]{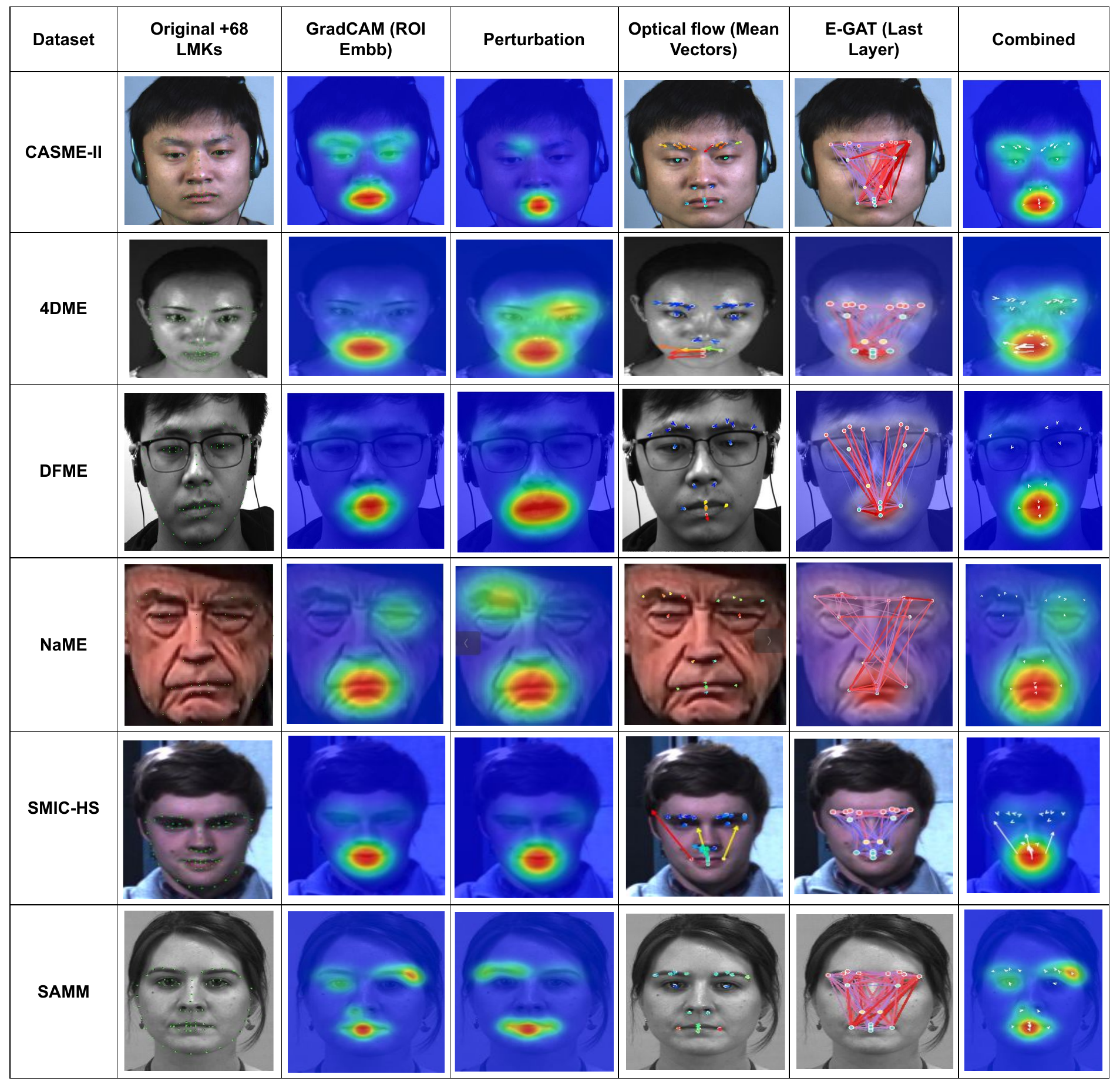}
    \caption{
Explainability analysis of the proposed STAG framework across multiple datasets using Grad-CAM, perturbation maps, optical flow, and E-GAT attention visualizations. Results are shown for positive classes (SMIC-HS and SAMM) and negative classes (remaining datasets), highlighting the model’s focus on discriminative facial action regions and motion patterns.}
    \label{fig:XAI}
\end{figure*}
Figure~\ref{fig:XAI} clearly validate the effectiveness of the pro-
posed STAG framework. The consistent and concentrated
activations across Grad-CAM, perturbation maps, optical
flow, and E-GAT demonstrate that the model reliably attends
to micro-expression relevant facial regions, particularly the
eyes and mouth, while suppressing irrelevant background
information. The sharper and more localized responses in
the combined model further confirm that integrating ap-
pearance, motion, and relational (AU-based graph) cues
leads to more discriminative and stable representations. This
alignment between visual explanations and known facial
action patterns provides strong evidence that STAG not only
achieves high recognition performance but also learns mean-
ingful and interpretable spatio-temporal features, thereby
validating its effectiveness for micro-expression recognition.

\begin{figure*}
    \centering
    \includegraphics[width=1.0\linewidth, keepaspectratio]{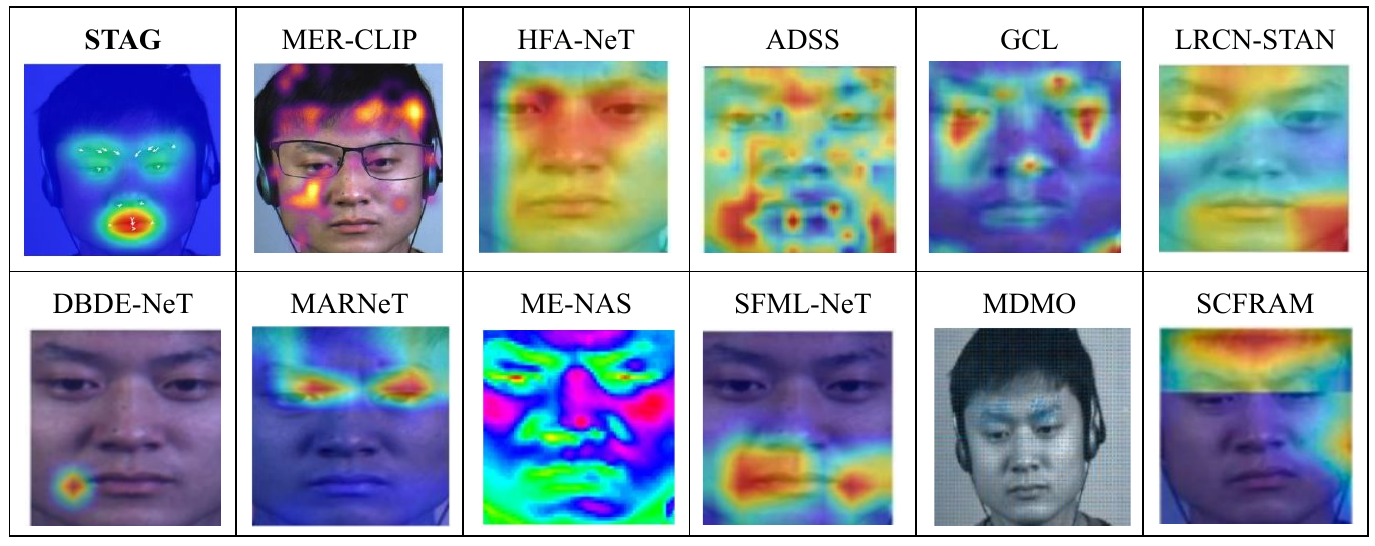}
     \caption{Qualitative comparison of heat-maps generated by different SOTA MER methods for the Figure~\ref{fig:XAI} illustration of the CASME-II dataset. 
    }
    \label{fig:Sota}
\end{figure*}
Figure~\ref{fig:Sota} illustrates the qualitative comparison of feature
attention maps produced by different SOTA MER
methods. Existing approaches such as MER-CLIP, HFA-
Net, ADSS, GCL, and SCFRAM often generate dispersed
or noisy activations across irrelevant facial regions, which
may weaken the representation of subtle micro-expression
cues. In contrast, the proposed STAG framework focuses
more precisely on action-unit-related regions, particularly
around the eyes and mouth, while suppressing background
interference.
\begin{figure*}[!t]
    \centering
    \includegraphics[width=1.0\linewidth, keepaspectratio]{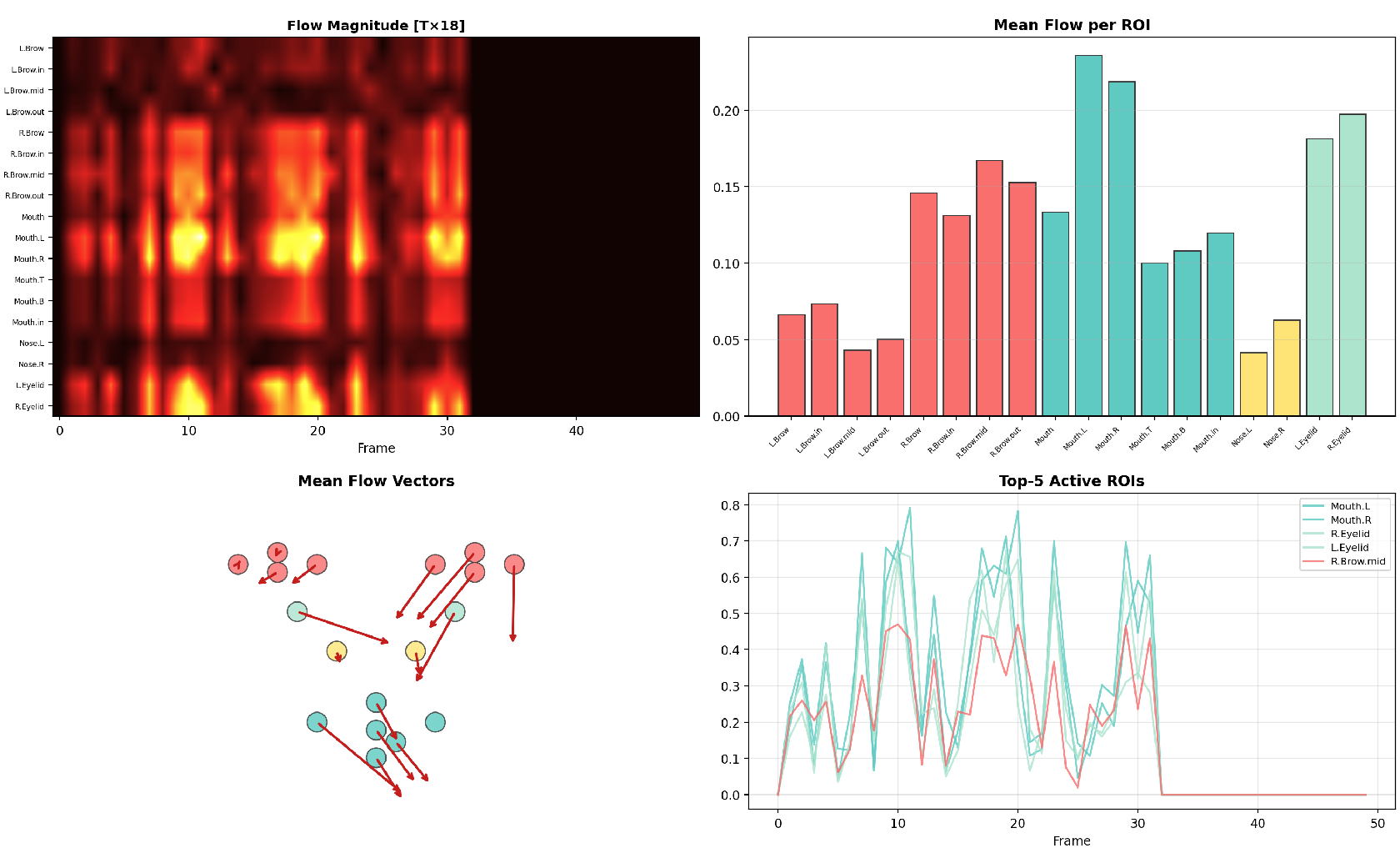}
    \caption{
    Visualization of ROI-based optical flow analysis for a micro-expression sample predicted as mentioned in the Figure~\ref{fig:XAI} for SAMM dataset. 
    The figure illustrates temporal optical-flow magnitudes across facial regions, average motion intensity per ROI, mean motion vectors, and the temporal evolution of the five most active facial regions.
    }
    \label{fig:3_optical_flow}
    \vspace{-15pt}
\end{figure*}

Figure~\ref{fig:3_optical_flow} provides an interpretable visualization of the
facial motion patterns captured by the proposed framework.
The heatmap (top-left) shows the temporal distribution of
optical-flow magnitudes across 18 facial ROIs, highlighting
localized motion primarily around the mouth, eyebrows, and
eyelids. The bar chart (top-right) summarizes the average
motion intensity of each ROI, revealing that the mouth and
eye regions contribute most strongly to the predicted ex-
pression. The mean flow vector plot (bottom-left) illustrates
the dominant direction and magnitude of facial movements,
while the temporal curves of the top-5 active ROIs (bottom-
right) show the evolution of motion intensity across frames.
Together, these visualizations demonstrate that the model
focuses on meaningful facial muscle activations associated
with the positive micro-expression.

\begin{figure*}
    \centering
    \includegraphics[width=1.0\linewidth, keepaspectratio]{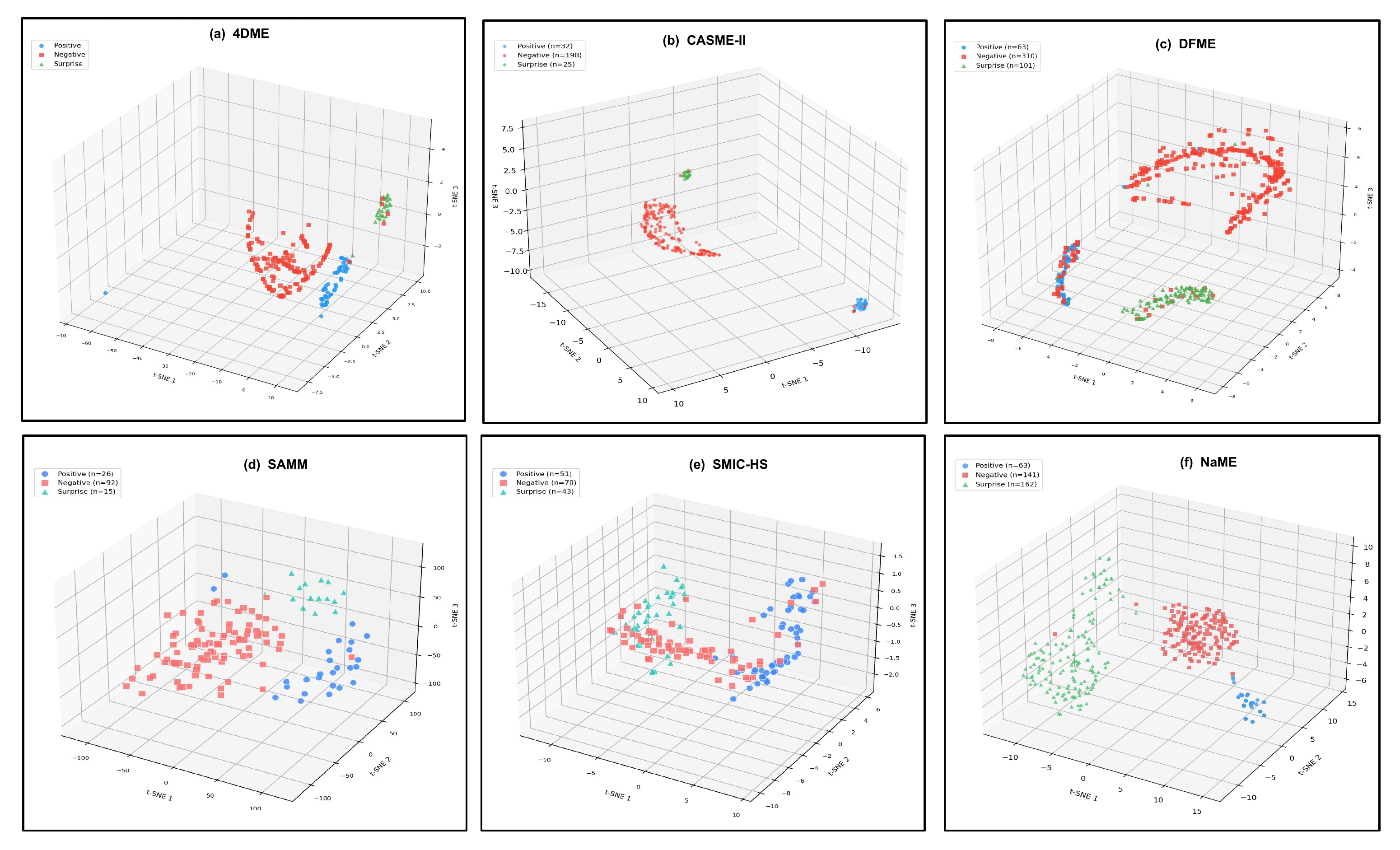}
    \caption{
    t-SNE visualization of learned feature embeddings.
    The proposed STAG method produces well-separated clusters for different emotion classes, indicating strong discriminative capability. Compared to baseline representations, STAG reduces class overlap and improves intra-class compactness, particularly for subtle expressions.
    }
    \label{fig:tsne}
    \vspace{-10pt}
\end{figure*}
\begin{figure*}
    \centering
    \includegraphics[width=1.0\linewidth, keepaspectratio]{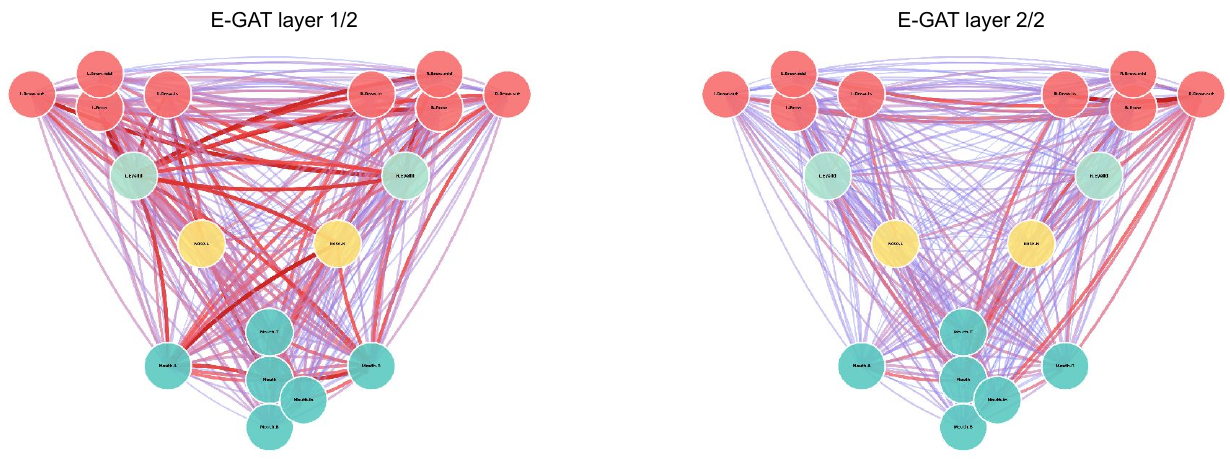}
    \caption{Spatial connection and message-passing visualization across the two sequential layers of the E-GAT. Node positions correspond structurally to facial ROIs, where edge colors and thicknesses signify the relative attention weights dynamically assigned to spatial dependencies during micro-expression processing of the Figure~\ref{fig:XAI} illustration of the SAMM dataset.}
    \label{fig:EGAT}
\end{figure*}

\begin{figure*}[!t]
    \centering
    \includegraphics[width=1.0\linewidth, height=0.3\linewidth]{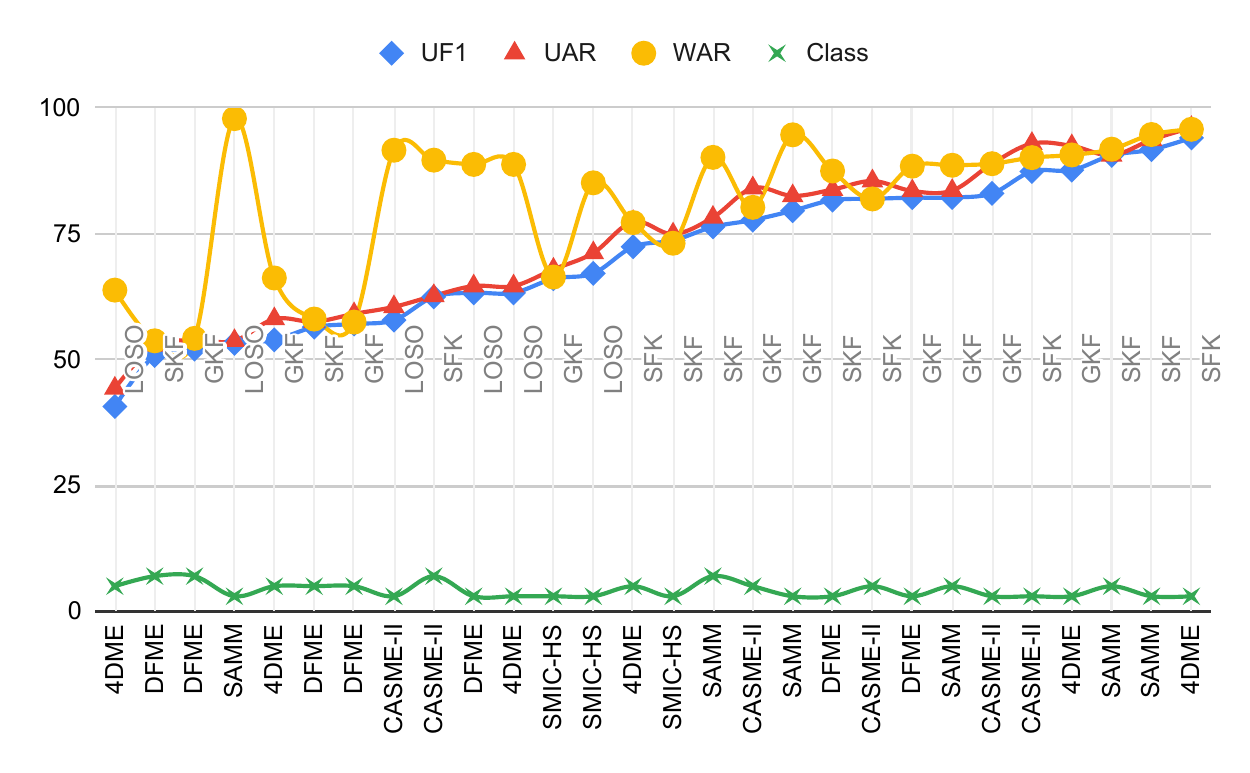}
    \caption{Performance comparison of the proposed STAG framework across multiple MER benchmark datasets under different evaluation protocols (LOSO, SKF, and GKF). The plot illustrates the variation of UF1, UAR, and WAR scores with respect to dataset complexity and class settings (3-class, 5-class, and 7-class protocols).}
    \label{fig:overall_performance}
    \vspace{-20pt}
\end{figure*}

\begin{figure*}[!t]
    \centering
    \includegraphics[width=1.0\linewidth, keepaspectratio]{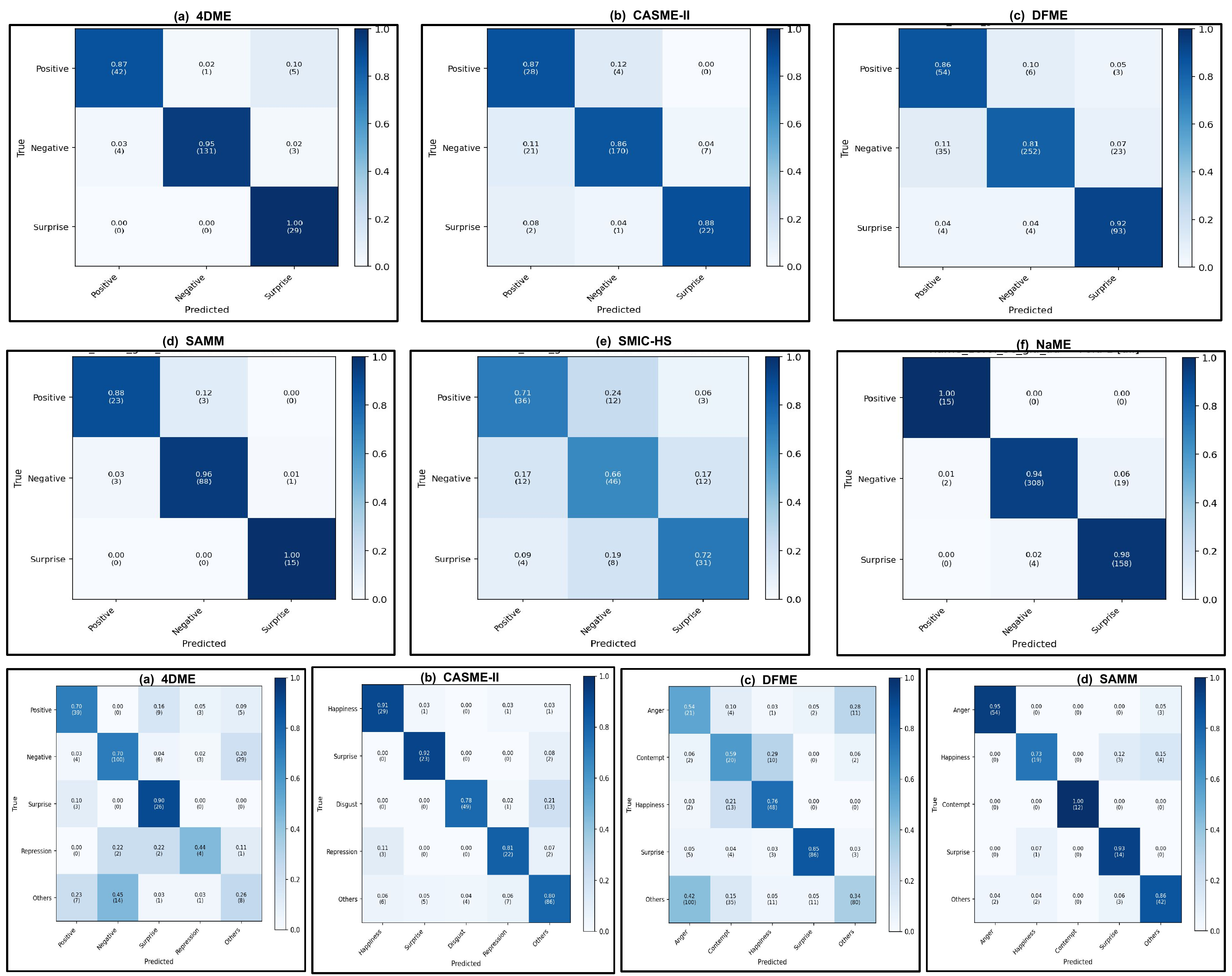}
    \caption{
Confusion matrices of STAG on six benchmark micro-expression datasets (4DME, CASME-II, DFME, SAMM, SMIC-HS, and NaME) for 3-class (top) and 5-class (bottom) classification tasks. Strong diagonal patterns indicate high recognition accuracy, while off-diagonal entries highlight confusion among visually similar and low-intensity micro-expressions. Despite increased difficulty in the 5-class setting, STAG maintains robust discriminative performance across datasets.
    }
    \label{fig:CM3_5}
\end{figure*}
Figure~\ref{fig:tsne} shows the 3D t-SNE feature embeddings for
the 4DME, CASME-II, DFME, SAMM, SMIC-HS, and
NaME datasets. Clear clustering of \textit{Positive}, \textit{Negative}, and \textit{Surprise} classes demonstrates the discriminative power of
the proposed framework. NaME and CASME-II exhibit the
most compact and well-separated clusters, while SMIC-HS
shows greater overlap due to subtle expression variations.
Moderate overlap is also observed in DFME and 4DME.
Overall, the results confirm effective feature learning and
class separability across diverse micro-expression datasets.

To evaluate the structural dependencies captured by the
spatial graph network, we visualize the learned attention dis-
tributions across the sequential layers of the E-GAT in Figure~\ref{fig:EGAT}. The node topologies are organized to reflect the physi-
cal anatomical distribution of the 18 facial Regions of Inter-
est (ROIs), grouping upper-facial features (e.g., eyebrows)
and lower-facial features (e.g., mouth regions) respectively.
In the first layer (E-GAT Layer 1/2), the network estab-
lishes dense, cross-regional attention maps, indicated by the
pronounced red and purple interconnected edges, capturing
broad, global synchronized movements across disparate fa-
cial muscle regions during a micro-expression event. Con-
versely, the second layer (E-GAT Layer 2/2) demonstrates
a clear structural refinement; the attention weights become
sparser and highly localized, focusing predominantly on
fine-grained, intra-region localized dynamics (such as lo-
cal eye or mouth micro-movements) and critical structural
boundaries. This layer-wise progression confirms that the E-
GAT effectively shifts from extracting global coarse-grained
facial context to isolating local, highly discriminative lo-
calized muscular variations necessary for precise micro-
expression classification.

Figure~\ref{fig:overall_performance} summarizes the performance of the proposed
framework across multiple datasets and evaluation protocols
(LOSO, SKF, and GKF) under 3, 5, and 7-class settings.
The model consistently achieves strong results, with the
best performance obtained on 4DME under the 3-class SKF
protocol (UF1 = 94.10\%, UAR = 96.13\%, WAR = 95.80\%).
SAMM and CASME II also show competitive performance,
confirming the effectiveness of the proposed graph-temporal
representation. Although performance decreases in the 5-
class and 7-class settings due to increased class complex-
ity and imbalance, the framework maintains stable results
across datasets. Overall, the findings demonstrate the ro-
bustness, generalization capability, and effectiveness of the
proposed approach for micro-expression recognition. 

Figure~\ref{fig:CM3_5} presents the confusion matrices of STAG
under 3-class and 5-class settings. The model achieves
strong recognition performance with high diagonal domi-
nance across datasets, particularly on NaME, SAMM, and
CASME-II. Although increased class complexity introduces
confusion in 4DME and DFME, especially among visually
resents the confusion matrices of STAG
under 3-class and 5-class settings. The model achieves
strong recognition performance with high diagonal domi-
nance across datasets, particularly on NaME, SAMM, and
CASME-II. Although increased class complexity introduces
confusion in 4DME and DFME, especially among visually

Overall, the results confirm that the proposed framework
effectively integrates graph attention learning, transformer-
based temporal modeling, and AU-guided spatial connec-
tivity to achieve robust and generalized micro-expression
recognition performance across multiple datasets and evalu-
ation settings.

\begin{table*}[!t]
\centering
\caption{List of symbols and notations used in the proposed STAG framework.}
\label{tab:symbols}
\begin{tabular}{cl|cl}
\hline
\textbf{Symbol} & \textbf{Description} &
\textbf{Symbol} & \textbf{Description} \\
\hline

$\mathcal{V}$ & Input micro-expression video &
$I_t$ & RGB frame at time step $t$ \\

$T$ & Number of temporal frames &
$H,W,C$ & Image height, width, and channels \\

$\mathcal{L}_t$ & Facial landmarks at frame $t$ &
$N_L$ & Number of facial landmarks (68) \\

$\mathbf{l}_t^{(i)}$ & Landmark coordinate &
$R$ & Number of facial ROIs \\

$\mathcal{R}_{t,r}$ & ROI region at frame $t$ &
$d_e$ & Inter-ocular distance \\

$g$ & Face scaling factor &
$\mathbf{c}_t$ & Facial center coordinate \\

$\mathbf{F}_{t,r}(p)$ & Optical flow vector &
$u_{t,r}(p),v_{t,r}(p)$ & Horizontal and vertical flow \\

$m_{t,r}(p)$ & Motion magnitude &
$\tau_{t,r}$ & Motion threshold \\

$\mathcal{P}_{t,r}$ & Dominant motion pixels &
$\mathbf{x}_{t,r}$ & Aggregated ROI motion vector \\

$\mathbf{X}$ & Optical flow tensor &
$\mathbf{h}_{t,r}$ & ROI embedding feature \\

$\mathbf{H}^{(0)}$ & Initial ROI embedding tensor &
$D_e$ & ROI embedding dimension \\

$D_h$ & MLP hidden dimension &
$B$ & Mini-batch size \\

$\mathcal{G}_t$ & Dynamic graph &
$\mathbf{A}^{(0)}$ & Initial adjacency matrix \\

$\hat{\mathbf{A}}^{(t)}$ & Dynamic adjacency matrix &
$L_g$ & Number of E-GAT layers \\

$K$ & Number of attention heads &
$d_h$ & Attention-head dimension \\

$e_{tijk}$ & Attention score &
$b_{ij}$ & Learnable edge bias \\

$\lambda$ & Temporal smoothing coefficient &
$\mathbf{E}^{(t)}$ & Aggregated attention matrix \\

$\omega_t$ & Temporal attention weight &
$\mathbf{G}$ & Graph token sequence \\

$\mathbf{f}_{graph}$ & Graph representation &
$D_g$ & Graph feature dimension \\

$\mathbf{H}_{flat}$ & Flattened ROI sequence &
$\mathbf{P}$ & Positional encoding \\

$L_t$ & Transformer layers &
$D_t$ & Transformer dimension \\

$\mathbf{T}$ & Transformer token sequence &
$\mathbf{f}_{trans}$ & Transformer representation \\

$\mathbf{G}^{*}$ & Cross-attended graph tokens &
$\mathbf{T}^{*}$ & Cross-attended transformer tokens \\

$D_f$ & Fusion projection dimension &
$\mathbf{f}_{fusion}$ & Fused representation \\

$N_{AU}$ & Number of Action Units &
$\mathbf{v}_{au}$ & Binary AU vector \\

$\mathbf{f}_{au}$ & AU embedding &
$D_{au}$ & AU embedding dimension \\

$\mathbf{f}_{final}$ & Final representation &
$\hat{\mathbf{y}}$ & Predicted logits \\

$\mathcal{Y}$ & Set of classes &
$C$ & Number of classes \\

$\alpha_c$ & Class balancing weight &
$\gamma_c$ & Focal-loss parameter \\

$\varepsilon_c$ & Label smoothing factor &
$\mathcal{L}$ & Training loss \\

$\theta$ & Trainable parameters &
$f_\theta(\cdot)$ & Proposed network \\

\hline
\end{tabular}
\end{table*}

\begin{table*}[!t]
\centering
\caption{Tensor dimensions used in the proposed STAG framework.}
\label{tab:tensor_dims}
\begin{tabular}{cl|cl|cl}
\hline
\textbf{Variable} & \textbf{Dimension} &
\textbf{Variable} & \textbf{Dimension} &
\textbf{Variable} & \textbf{Dimension} \\
\hline

$\mathbf{X}$ &
$\mathbb{R}^{T\times R\times 2}$ &
$\mathbf{x}_{t,r}$ &
$\mathbb{R}^{2}$ &
$\mathbf{F}_{t,r}(p)$ &
$\mathbb{R}^{2}$ \\

$\mathbf{h}_{t,r}$ &
$\mathbb{R}^{D_e}$ &
$\mathbf{H}^{(0)}$ &
$\mathbb{R}^{T\times R\times D_e}$ &
$\mathbf{A}^{(0)}$ &
$\mathbb{R}^{R\times R}$ \\

$\hat{\mathbf{A}}^{(t)}$ &
$\mathbb{R}^{R\times R}$ &
$\mathbf{Q}_{tik},\mathbf{K}_{tjk}$ &
$\mathbb{R}^{d_h}$ &
$\mathbf{E}^{(t)}$ &
$\mathbb{R}^{R\times R}$ \\

$\mathbf{H}^{(\ell)}_t$ &
$\mathbb{R}^{R\times D_e}$ &
$\mathbf{V}_t$ &
$\mathbb{R}^{R\times D_e}$ &
$\mathbf{G}$ &
$\mathbb{R}^{R\times D_e}$ \\

$\mathbf{g}_r$ &
$\mathbb{R}^{D_e}$ &
$\mathbf{z}_t$ &
$\mathbb{R}^{R D_e}$ &
$\mathbf{z}$ &
$\mathbb{R}^{R D_e}$ \\

$\mathbf{f}_{graph}$ &
$\mathbb{R}^{D_g}$ &
$\mathbf{H}_{flat}$ &
$\mathbb{R}^{T\times (R D_e)}$ &
$\mathbf{P}$ &
$\mathbb{R}^{(T+1)\times D_t}$ \\

$\hat{\mathbf{H}}^{(0)}$ &
$\mathbb{R}^{(T+1)\times D_t}$ &
$\mathbf{T}$ &
$\mathbb{R}^{T\times D_t}$ &
$\mathbf{f}_{trans}$ &
$\mathbb{R}^{D_t}$ \\

$\mathbf{G}_f$ &
$\mathbb{R}^{R\times D_f}$ &
$\mathbf{T}_f$ &
$\mathbb{R}^{T\times D_f}$ &
$\mathbf{G}^{*}$ &
$\mathbb{R}^{R\times D_f}$ \\

$\mathbf{T}^{*}$ &
$\mathbb{R}^{T\times D_f}$ &
$\mathbf{f}_g,\mathbf{f}_t$ &
$\mathbb{R}^{D_f}$ &
$\mathbf{f}_{fusion}$ &
$\mathbb{R}^{D_g+D_t}$ \\

$\mathbf{v}_{au}$ &
$\{0,1\}^{N_{AU}}$ &
$\mathbf{f}_{au}$ &
$\mathbb{R}^{D_{au}}$ &
$\mathbf{f}_{final}$ &
$\mathbb{R}^{D_g+D_t+D_{au}}$ \\

$\hat{\mathbf{y}}$ &
$\mathbb{R}^{C}$ &
$\tilde{\mathbf{y}}$ &
$\mathbb{R}^{C}$ &
$p_{ic}$ &
$\mathbb{R}$ \\

\hline
\end{tabular}
\end{table*}

\section{Conclusion}\label{sec:Con}
This paper proposed STAG, a unified spatial–temporal
framework for micro-expression recognition that effectively
models subtle facial dynamics. The framework integrates
AU-guided dynamic ROI connectivity, graph attention-
based spatial reasoning, and transformer-based temporal
modeling to jointly capture fine-grained appearance, motion,
and relational cues. In addition, a cross-attention mechanism
is introduced to enhance interaction between spatial and tem-
poral representations, leading to more discriminative feature
learning. Extensive experiments on six benchmark datasets
under LOSO, group K-fold, and stratified K-fold protocols
demonstrate that STAG consistently achieves competitive
and often superior performance compared to existing state-of-the-art methods. Further cross-dataset evaluations con-
firm its strong generalization ability in unseen domain
settings, while ablation and explainability analyses validate
the contribution of each component and the interpretability
of the learned representations. Overall, the results indicate
that dynamic AU-guided relational modeling combined
with spatio-temporal attention is highly effective for micro-expression recognition. The proposed STAG framework
provides a robust and interpretable solution for XAI-based, advancing the
reliability and generalization of micro-expression analysis in
real-world scenarios. 
Future work
will investigate explainable vision-language graph learn-
ing frameworks that align evolving ROI–AU interactions
with textual facial behavior descriptions, improving both
interpretability and generalization across diverse micro-
expression datasets.

\section{Acknowledgment}
The authors express their profound appreciation to the Vision Intelligence and Machine Learning (VIML) Group for their important help, guidance, and collaborative attitude. Their proficiency and support have significantly enhanced the success of this endeavor. Gratitude is extended to Dr. Dinesh Singh for his guidance, unwavering support, and for facilitating access to high-performance GPUs and edge computing devices. The authors express their gratitude for the utilization of CHEETAH, a GPU-based computational facility established under research grant No. IITM/SG/DIS-ROS-SPA/111 at the Indian Institute of Technology Mandi, Department of Higher Education, Ministry of Education, Government of India, to fulfill the computational needs of this research endeavor. We acknowledge the utilization of Grammarly (Grammarly Inc.) for rectifying grammatical inaccuracies and improving the readability of this paper.

\bibliographystyle{cas-model2-names}
\bibliography{main}

\bio{images/photo}
Miss. Nandani Sharma is a doctoral candidate at IIT Mandi in the SCEE department, working under the supervision of Dr. Dinesh Singh. She earned her B.Tech. (Hons.) and M.Tech. in Computer Science and Engineering from Dr. A.P.J. Abdul Kalam Technical University (APJAKTU), where she was awarded the Gold Medal for securing the top position in her branch across Uttar Pradesh, India, under the supervision of Dr. Peeyush Kumar Pathak. Her work contributes to advancements in human-computer interaction, affective computing, and medical diagnostics. As a visual intelligence and machine learning (VIML) research group member at IIT Mandi, she contributes to computer vision, face analysis, deep learning, human-computer interaction, surveillance video analytics.
\endbio
\bio{images/1000326552}
Mr. Varun Sharma is a B.Tech. student in the Department of Electronics and Communication Engineering at the Indian Institute of Information Technology (IIIT) Bhagalpur. His research interests include artificial intelligence, machine learning, computer vision, representation learning, large language models, and generative AI systems. His work focuses on multimodal AI, image and video analysis, intelligent virtual assistants, and human-computer interaction. He is actively exploring modern AI technologies and their applications in interactive and creative computing systems, with particular interest in AI-driven research, visual intelligence, and advanced deep learning methodologies for real-world intelligent systems.
\endbio
\bio{images/DrDineshSingh}
Dr. Dinesh Singh is an Assistant Professor in the School of Computing and Electrical Engineering, Indian Institute of Technology Mandi. Prior to joining IIT Mandi, He worked in the High-dimensional Statistical Modeling Team as a Postdoctoral Researcher at the RIKEN Center for Advanced Intelligence Project (AIP), Kyoto University Office, Japan. He completed his Ph.D. degree in Computer Science and Engineering from IIT Hyderabad on Scalable and Distributed Methods for Large-scale Visual Computing. He received his M.Tech degree in Computer Engineering from NIT Surat on machine-learning approaches for network anomaly and intrusion detection in the domain of cybersecurity and cloud security. His research interests include machine learning, big data analytics, visual computing, and cloud computing.
\endbio

\end{document}